\def\EVOSTEXTENDED{}
\documentclass[sigconf]{acmart}
\AtBeginDocument{%
  }

\usepackage{algorithm}
\usepackage{algorithmic}
\usepackage{textcomp}
\usepackage{booktabs}       %
\usepackage{amsfonts}       %
\usepackage{hyphenat}       %
\usepackage{amsmath}        %
\usepackage{nicefrac}       %
\usepackage{microtype}      %
\usepackage{xcolor}         %
\usepackage{colortbl}       %
\usepackage{graphicx}
\usepackage{makecell}
\usepackage{multirow}
\usepackage{multicol}
\usepackage{booktabs,multirow}
\usepackage{array} %
\usepackage{scalefnt}
\usepackage{subcaption}
\usepackage{float}
\usepackage{cleveref}

\usepackage{placeins}
\usepackage{newfloat}
\usepackage{listings}
\usepackage{xcolor}
\definecolor{myblue}{HTML}{1D7AB8}
\definecolor{myred}{HTML}{E63946}
\definecolor{obsblue}{HTML}{1F4E79}   %
\definecolor{sizewine}{HTML}{85144b}  %
\usepackage{caption} %
\usepackage{xspace}
\newcommand{\ourmethod}{EvoST\xspace}
\newif\ifextended
\ifdefined\EVOSTEXTENDED\extendedtrue\fi
\newcommand{\extref}[2]{\ifextended#1\else#2\fi}

\setcopyright{none}
\copyrightyear{2027}
\acmYear{2027}
\acmDOI{}

\begin{document}

\title[EvoST: Adapting Traffic Forecasters]{EvoST: Adapting Traffic Forecasters\texorpdfstring{\\}{ }to Evolving Sensor Graphs}

\author{Du Yin}
\email{du.yin@unsw.edu.au}
\affiliation{%
  \institution{University of New South Wales}
  \city{Sydney} \state{NSW} \country{Australia}
}
\affiliation{%
  \institution{Hong Kong University of Science and Technology (Guangzhou)}
  \country{China}
}

\author{Xiachong Lin}
\email{dawn.lin@student.unsw.edu.au}
\affiliation{%
  \institution{University of New South Wales}
  \city{Sydney} \state{NSW} \country{Australia}
}

\author{Yue Tan}
\email{yue.tan@griffith.edu.au}
\affiliation{%
  \institution{Griffith University}
  \city{Brisbane} \state{Queensland} \country{Australia}
}

\author{Ruiyi Yang}
\email{ruiyi.yang@student.unsw.edu.au}
\affiliation{%
  \institution{University of New South Wales}
  \city{Sydney} \state{NSW} \country{Australia}
}

\author{Jinliang Deng}
\email{dengjinliang@ust.hk}
\affiliation{%
  \institution{The Hong Kong University of Science and Technology}
  \country{Hong Kong}
}

\author{Estrid He}
\email{estrid.he@rmit.edu.au}
\affiliation{%
  \institution{RMIT University}
  \city{Melbourne} \state{VIC} \country{Australia}
}

\author{Hao Xue}
\email{haoxue@hkust-gz.edu.cn}
\affiliation{%
  \institution{Hong Kong University of Science and Technology (Guangzhou)}
  \country{China}
}

\author{Flora Salim}
\email{flora.salim@unsw.edu.au}
\affiliation{%
  \institution{University of New South Wales}
  \city{Sydney} \state{NSW} \country{Australia}
}

\renewcommand{\shortauthors}{Yin et al.}

\begin{abstract}

Web-based navigation and traffic information services depend on accurate forecasts as sensor networks evolve. Controlled comparisons on checkpoints reused years later show that updating only shared weights or adding a low-rank node prompt leaves substantial degradation. In the evaluated architectures, node identities are an effective maintenance target: refreshing them recovers most of this degradation while the shared spatio-temporal weights are reused. We propose EvoST, a framework that maintains traffic forecasters by refreshing node identities while keeping shared weights fixed. EvoST updates existing identities using recently released labels and initializes new sensors with identities borrowed from behaviorally similar established sensors. Once sufficient delayed labels arrive, it jointly refits all sensor identities and recalibrates their outputs. A per-sensor recursive least-squares calibrator corrects forecasts as labels become available. Across twelve PeMS region-year settings, identity refresh reduces the relative MAE gap over the one-hour forecast horizon between five-year-old STAEFormer checkpoints and models trained in the target year from 17.7\% to 2.5\%, with all compared forecasters calibrated, and its MAE is 0.7\% above that of full-parameter fine-tuning on the same data. In Graph WaveNet, whose adaptive adjacency is generated from node embeddings, identity refresh recovers 88\% of the five-year MAE gap to calibrated target-year models. Similar recovery holds for STAEFormer checkpoints up to fifteen years old, while refreshing STID recovers 67\% of its five-year MAE gap to calibrated target-year counterparts. On an annual benchmark covering nine EvoXXLTraffic districts and Transport for NSW, EvoST has the lowest average MAE, RMSE, and MAPE among the reported non-foundation methods on all ten networks, and calibration accounts for most of its Avg-MAE reduction from the matched frozen checkpoints.

Our code and an extended version of this paper are available at \url{https://github.com/lixus7/EvoST}.

\end{abstract}



\maketitle

\section{Introduction}
Traffic forecasting is vital for building smart cities and intelligent transportation systems, helping alleviate congestion and improve mobility by supporting efficient traffic management and route planning~\cite{chen2025expand,shao2022spatial,wu2019graph, yu2018spatio, zhang2025strap,zhou2026feddis,shao2025hyperd,zhang2023autostl,zhang2023mlpst,zhang2023promptst,ma2023rethinking}. 
Usually, existing traffic forecasting studies are built upon fixed sensor graphs, where nodes represent traffic sensors deployed at specific locations and edges characterize the spatial connectivity of road networks. 
\extref{To model complex spatio-temporal traffic dynamics and accurately predict future traffic states, e.g., traffic flow, speed, and demand, various data-driven methods have been developed, typically following a workflow of traffic data collection, graph construction, spatio-temporal modeling, and future-state prediction. 
This pipeline has substantially broadened the range of downstream applications that traffic forecasting can support, such as congestion management, route planning, and traffic control, etc.}{To model complex spatio-temporal traffic dynamics and accurately predict future traffic states, e.g., traffic flow, speed, and demand, various data-driven methods have been developed, typically following a workflow of traffic data collection, graph construction, spatio-temporal modeling, and future-state prediction.}

\begin{figure}[!h]
    \centering
    \includegraphics[width=\linewidth]{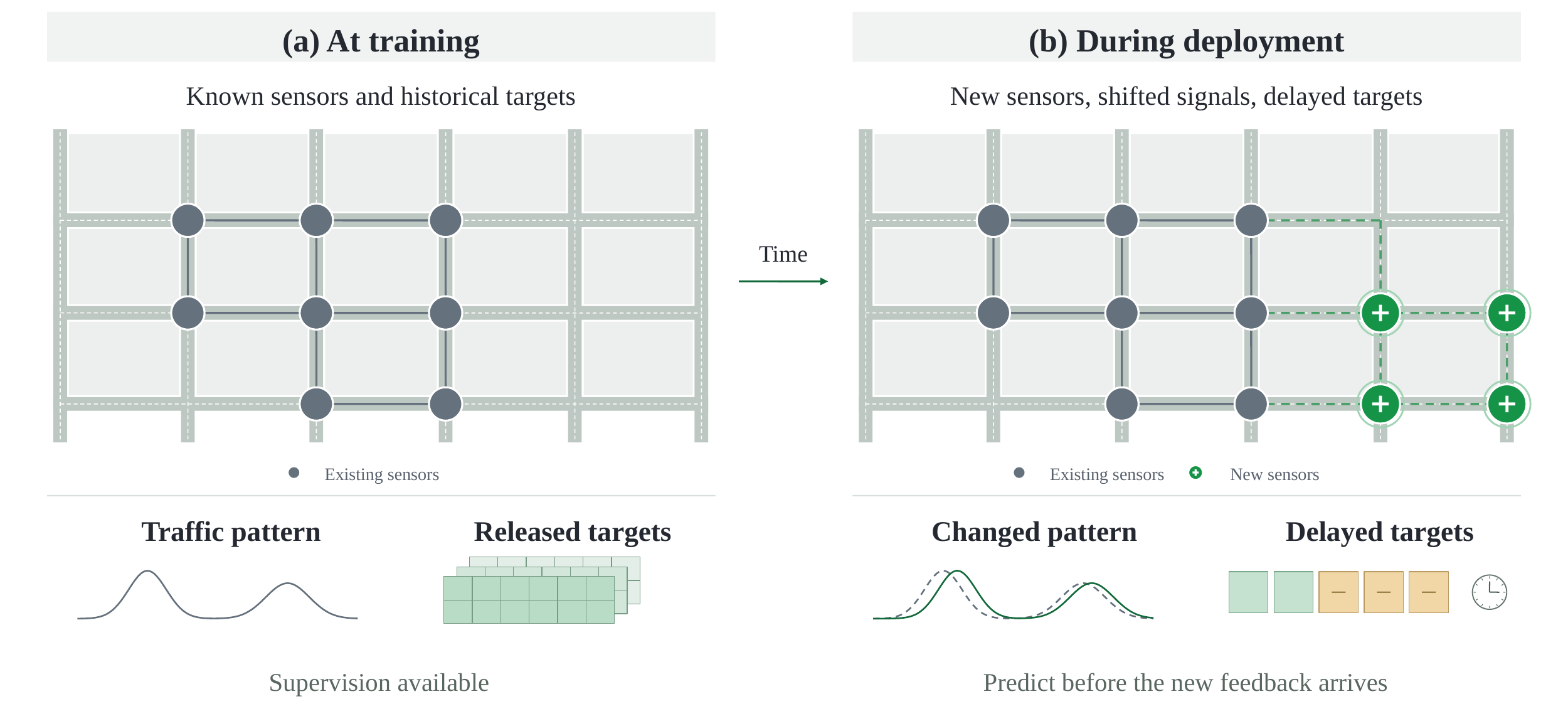}
    \caption{Traffic forecasting on evolving sensor networks. (a) At training time, historical observations and targets are available for known sensors. (b) During deployment, new sensors join the network and traffic patterns change. Forecasts must be issued before their delayed targets are released. The traffic curves are schematic.}
    \label{fig:a2tta_intro}
    \Description{Two panels show a road network before and after new sensors are added. Existing sensors are gray and new sensors are green. Beneath the networks, schematic traffic curves and target blocks contrast available training supervision with changing signals and delayed feedback during deployment.}
\end{figure}

Traditional methods generally assume a static traffic network~\cite{guo2019attention, jiang2023pdformer, shao2022decoupled, jiang2021dl}, following a train-once-and-deploy paradigm in which the trained model is expected to remain effective in a long period. 
Unfortunately, this assumption rarely holds in practice, as real-world traffic sensor networks continuously evolve over time. 
On the one hand, road networks undergo continuous construction and renewal, leading to the deployment of new sensors and the removal of outdated ones. Meanwhile, ongoing urbanization reshapes road layouts and connectivity, further altering the underlying sensor graph. 
On the other hand, human mobility patterns also evolve over time due to changes in commuting behavior, population distribution, and transportation policies, resulting in shifts in traffic dynamics even when the road topology remains unchanged. 
In these cases, conventional train-once-and-deploy forecasting models may suffer from substantial performance degradation, as both the underlying sensor graph and traffic dynamics progressively shift away from those observed during training. 
To handle the above dynamic changes, a straightforward solution is to regularly train the model. However, retraining can be expensive and may overwrite useful knowledge with short-lived patterns. 
Under this circumstance, a promising alternative is to adapt a pretrained model with data observed during deployment, as in test-time adaptation (TTA)~\cite{guo2024online, wangtent}, rather than retraining it from scratch. 
Such adaptation reuses the knowledge encoded in the pretrained backbone, but it leaves open which of the backbone's parameters should be updated.

Adapting a checkpoint to an evolving sensor network raises three difficulties.
First, it is unclear which parameters become outdated as the network evolves. Existing methods update lightweight output modules~\cite{guo2024online}, node prompts~\cite{chen2025expand}, or the whole model with replay~\cite{ijcai2021p0498}.
Wang et al.~\cite{wang2024inflexibility} study cross-year failure of adaptive embeddings and show that fine-tuning only these embeddings on a small target-year subset improves accuracy one year later on an unchanged sensor set. We ask whether such updates suffice for checkpoints reused five to fifteen years later, on networks that gain and lose sensors and receive labels with delay, and how they compare with matched updates of other parameter groups.
Second, newly deployed sensors have short histories, which make reliable fitting of their sensor-specific parameters difficult at arrival.
Third, as Figure~\ref{fig:a2tta_intro} illustrates, a target becomes available only after its forecast horizon has elapsed, so any correction must rely on delayed feedback.

We therefore first ask which parameters need maintenance.
We reuse STAEFormer checkpoints trained five years before the target year on twelve PeMS region-year settings. With all forecasters calibrated, their MAE over the full one-hour forecast horizon is 17.7\% higher than that of checkpoints trained in the target year.
Using the same recent windows, tuning only the shared spatio-temporal weights, the output head, or an EAC-style input prompt recovers much less of this gap than refreshing only the node identities, which recovers 86\% of it, with MAE 0.7\% above that of full fine-tuning (Section~\ref{sec:which_params}).
In the evaluated architectures, node identities are thus an effective maintenance target: refreshing them recovers most of the measured degradation, and the shared weights remain largely reusable with calibration.

Building on this diagnosis, we propose \ourmethod, which maintains a deployed forecaster by refreshing node identities while keeping its shared spatio-temporal weights fixed (Figure~\ref{fig:a2tta_framework}).
\ourmethod refits the identities of sensors with recent labeled history. For backbones that generate adaptive adjacency from node embeddings, such as Graph WaveNet~\cite{wu2019graph}, this also updates the learned graph.
A new sensor starts from the mean identity of behaviorally similar established sensors and joins a refit of all identities once enough delayed labels have been released.
Between fitting stages, a per-sensor recursive least-squares calibrator corrects forecasts using only targets whose complete horizon has elapsed.

Our main contributions are summarized as follows:
\begin{itemize}
    \item We study the maintenance of traffic forecasters whose checkpoints are reused five to fifteen years after training, on networks that gain and lose sensors and with labels that arrive with delay. Using matched recent data and calibration within each setting, we compare updating node identities, shared weights, the output head, a node prompt, or the full model of five-year-old STAEFormer checkpoints, and test identity refresh on two more architectures and on checkpoints up to fifteen years old. Node identities are an effective target, and the shared spatio-temporal weights remain largely reusable.

    \item We propose \ourmethod, a maintenance procedure that refreshes node identities with the shared backbone frozen, lets sensors the checkpoint has never seen borrow identities from behaviorally similar established sensors until a joint refit, and corrects forecasts with per-sensor recursive least squares under the same delayed-feedback rule.

    \item With all compared forecasters calibrated, identity refresh recovers 86\% of the MAE gap over the one-hour horizon between five-year-old and target-year STAEFormer checkpoints on twelve PeMS region-year settings. Similar recovery holds for STAEFormer checkpoints up to fifteen years old. On sixteen settings, STID and Graph WaveNet recover 67\% and 88\% of their respective five-year gaps. On an annual benchmark covering nine EvoXXLTraffic districts and TFNSW with a separate checkpoint for each year, \ourmethod has the lowest average MAE, RMSE, and MAPE of the reported non-foundation methods. On average across these networks, calibration accounts for most of the Avg-MAE reduction from the matched frozen checkpoints.
\end{itemize}

\section{Related Work}
\label{sec:related_work}

Traffic forecasting has developed from statistical time-series models~\cite{box2015time,williams2003modeling} to neural models of temporal and spatial dependencies.
T-GCN, STGCN, and DCRNN combine temporal modeling with graph convolutions or diffusion over sensor graphs~\cite{zhao2019t,yu2018spatio,li2018diffusion}.
Graph WaveNet learns adaptive adjacency, while GMAN and STTN use attention to model spatial dependencies~\cite{wu2019graph,zheng2020gman,xu2020spatial}.
STID and STAEFormer use learnable node and spatio-temporal embeddings within MLP and Transformer backbones, respectively~\cite{shao2022spatial,liu2023spatio}.
These models are generally evaluated with the same sensors in training and testing.
Wang et al.~\cite{wang2024inflexibility} study cross-year embedding failure and PCA-based transfer across graphs. Their one-year comparison on matched sensors includes embedding-only fine-tuning on a small target-year subset.

Continual traffic forecasting methods address changing networks through replay, pattern memories, and selective updates~\cite{ijcai2021p0498,wang2023pattern,wang2023knowledge}.
EAC keeps the forecasting backbone frozen and adapts a pool of node prompts through expansion and low-rank compression~\cite{chen2025expand}.
STRAP retrieves historical spatio-temporal patterns to augment predictions~\cite{zhang2025strap}, while ST-TTC updates a spectral output calibrator using delayed labels~\cite{chen2025learning}.
We instead reuse checkpoints five to fifteen years after training on networks that gain and lose sensors, with labels released only after the forecast horizon, and compare node identities, shared weights, the output head, a node prompt, and the full model under the same data and calibration.
EvoST then refreshes existing node representations, including those that define learned adjacency, and lets never-seen sensors borrow identities until a joint refit, with the shared backbone frozen.
\extref{Further related work is discussed in Appendix~\ref{app:extended_related_work}.}{}

\section{Dataset and Empirical Motivation}
\subsection{EvoXXLTraffic Construction}

EvoXXLTraffic organizes five-minute records from the California Department of
Transportation Performance Measurement System (PeMS)
~\cite{chen2001freeway,yin2026from} into yearly graph snapshots for nine
districts (D03--D08 and D10--D12). For each district, $\mathcal V_y$
contains sensors observed on the first available day in the initial year
and on January~1 in subsequent years. Persistent IDs align the flow
matrices and identify additions and removals between yearly snapshots.
Sensor coordinates define a spatial proximity graph $\mathbf A_y$ through
Haversine distances and a thresholded Gaussian kernel ($\epsilon=0.1$).
TFNSW is a long-span external test network.

\subsection{Sensor Network Growth}

The sensor cohorts extend and densify coverage along road corridors, with
sensor counts increasing by up to approximately $95\times$.
Appendix~\ref{app:structural_evolution} maps the cohorts, gives the
district-level scale, and shows non-monotonic changes in average degree and
graph density, motivating adaptation to changing sensor sets and spatial
connectivity.

\section{Problem Formulation}
Following EvoXXLTraffic~\cite{yin2026from,yin2025xxltraffic}, we model each district as yearly
graph snapshots whose sensor set, adjacency matrix, and traffic observations
may change.

\subsection{Evolving Sensor Graph}
For district $d$ and year $y$, let $\mathbf X_y^d$, $\mathbf A_y^d$,
and $\mathcal V_y^d$ denote the traffic tensor, adjacency matrix, and
active sensor set:
\begin{equation}
    \mathcal D_{\mathrm{evo}}^d
    =\left\{\left(\mathbf X_y^d,\mathbf A_y^d,\mathcal V_y^d\right)
      \right\}_{y=y_0}^{y_1}.
\end{equation}
Here, $\mathbf X_y^d\in\mathbb R^{T_y^d\times N_y^d\times C}$,
$N_y^d=|\mathcal V_y^d|$, and $T_y^d$ and $C$ are the numbers of time
steps and channels. Both $\mathcal V_y^d$ and $\mathbf A_y^d$ may change
between years. Sensors newly active relative to the preceding year form
\begin{equation}
    \Delta\mathcal V_y^d=\mathcal V_y^d\setminus\mathcal V_{y-1}^d.
\end{equation}
The intersection contains retained sensors, while
$\mathcal V_{y-1}^d\setminus\mathcal V_y^d$ contains inactive ones.
This annual expansion cohort appears in year $y$'s training, validation,
and test partitions.

We omit the district superscript below. Let $\mathcal V_{\mathrm{ckpt}}$
contain the sensor IDs represented in the reused checkpoint.
The active sensors with checkpoint identities are
$\mathcal K=\mathcal V_y\cap\mathcal V_{\mathrm{ckpt}}$. Those absent
from the checkpoint are $\mathcal U=\mathcal V_y\setminus\mathcal K$.
Thus, $\mathcal U$ depends on the checkpoint year and can differ from
the annual expansion cohort $\Delta\mathcal V_y$.
In the arrival evaluation, each sensor in $\mathcal U$ has a specified
arrival time. Its identity initialization, donor matching, and
sensor-specific fitting use observations from that time onward.

\subsection{Forecasting Task}
We focus on univariate traffic flow ($C=1$). Within snapshot $y$, let
$\mathbf s_t\in\mathbb R^{N_y}$ denote the traffic state at time $t$.
The input and target windows are
\begin{equation}
    \begin{aligned}
    \mathbf X_t
    &= [\mathbf s_{t-L+1},\ldots,\mathbf s_t]
       \in\mathbb R^{N_y\times L},\\
    \mathbf Y_t
    &= [\mathbf s_{t+1},\ldots,\mathbf s_{t+H}]
       \in\mathbb R^{N_y\times H}.
    \end{aligned}
\end{equation}
We process one chronological window at a time. Before predicting at $t$,
record $i$ can supply feedback only after its complete target window
has elapsed, that is, $i+H\leq t$.

\paragraph{Checkpoint maintenance.}
We reuse a checkpoint trained in an earlier year, keeping its shared
spatio-temporal weights and normalization statistics fixed.
Available labeled history supports node identity fitting and output
calibration. Subsequent feedback updates the calibrators under the
complete-horizon release rule. For incoming sensors with short histories,
EvoST borrows identities before a scheduled refit.
The task is to maintain forecasting accuracy on the current sensor network
through these updates.

\section{Methodology}
\label{headings}

\begin{figure*}[t]
    \centering
    \includegraphics[width=0.95\linewidth]{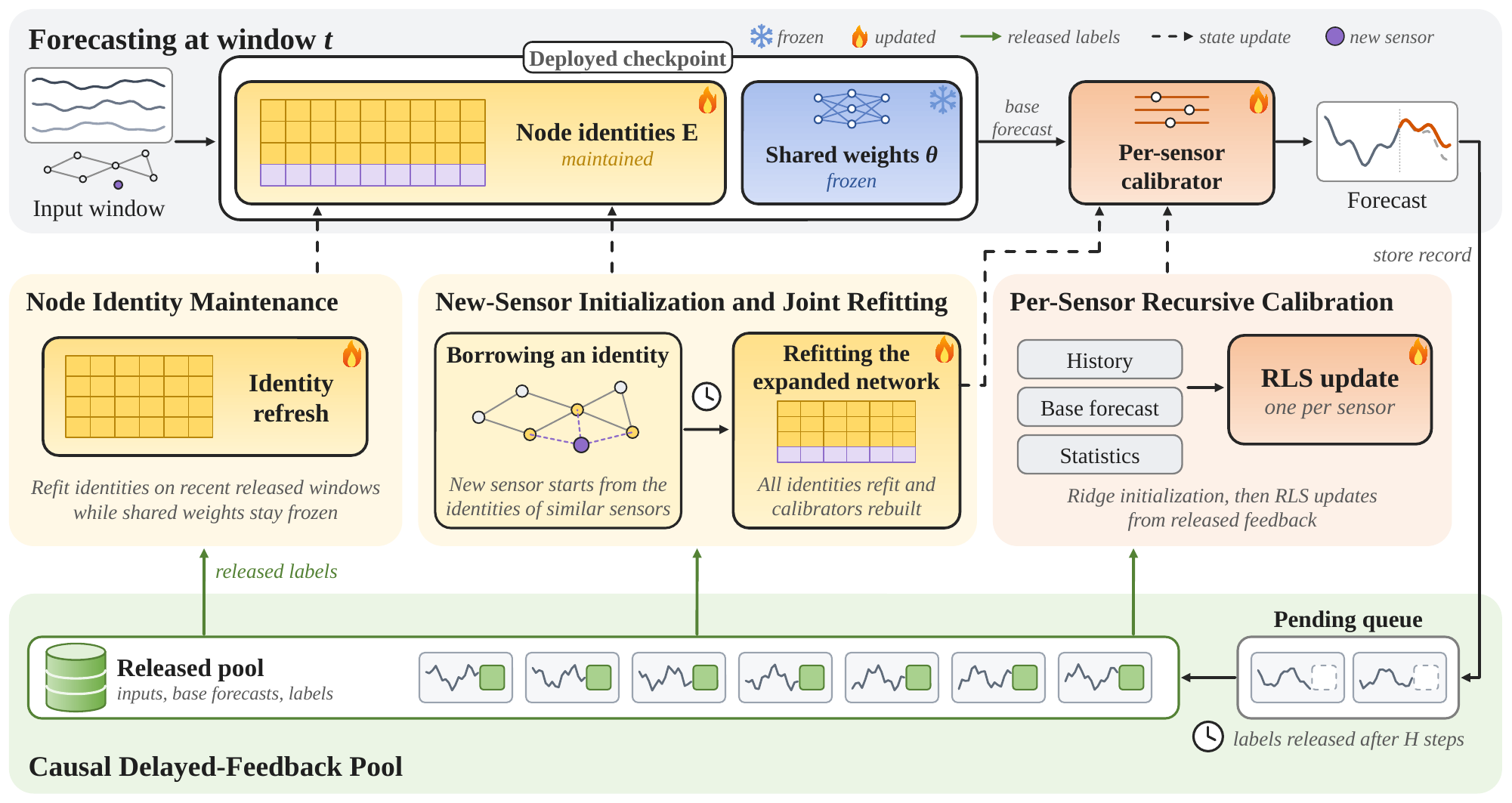}
    \caption{Overview of EvoST. At each window, the deployed checkpoint
    combines maintained node identities $\mathbf E$ with frozen shared
    weights $\boldsymbol\theta$, and per-sensor calibrators correct the
    base forecast. Each forecast record stores its input and base forecast in a pending
    queue until its labels are released $H$ steps later.
    Released labels drive three maintenance modules
    (Sections~\ref{sec:identity_maintenance}, \ref{sec:arrival_refit},
    and~\ref{sec:recursive_calibration}), each of which updates the state
    above it. Blue marks frozen weights, amber node identities, orange
    calibration, green released labels, and purple new sensors.
    Snowflakes mark frozen parts, flames mark updated parts, and dashed
    arrows mark state updates.}
    \label{fig:a2tta_framework}
    \Description{Three horizontal bands. The top band shows forecasting at
    one window. An input window enters a deployed checkpoint with maintained node
    identities and frozen shared weights. Adjacency is supplied when used
    by the backbone. The base forecast passes through a per-sensor
    calibrator to the final forecast. The middle band shows three maintenance modules: node
    identity maintenance, new-sensor initialization and joint refitting,
    and per-sensor recursive calibration. Dashed arrows show each module
    updating the identities or the calibrator above it. The bottom band
    shows the causal delayed-feedback pool. Records containing inputs and issue-time base forecasts wait in a
    pending queue and enter the released pool once their complete-horizon
    labels become available. Green arrows carry the released labels up to
    the three modules.}
\end{figure*}

\subsection{Overview}
Figure~\ref{fig:a2tta_framework} illustrates EvoST.
Given the current sensor network and input $\mathbf X_t$,
the forecaster produces a base prediction using
frozen shared weights and maintained node identities.
When recent labeled history is available for all active sensors, EvoST
fits their identities before deployment. With restricted arrival history,
established sensors are first refreshed on their own network, while
incoming sensors borrow identities from established sensors with similar
observations. A later refit updates all active identities, followed by
reinitialization of the output calibrators.
Between fitting stages, per-sensor recursive least squares (RLS)
updates the forecast correction using released feedback.

\subsection{Node Identity Maintenance}
\label{sec:identity_maintenance}

\paragraph{Parameters.}
We separate the checkpoint into shared parameters $\boldsymbol\theta$
and node identities $\mathbf E$:
\begin{equation}
    \widehat{\mathbf Y}^{\mathrm{base}}_t
    =f_{\boldsymbol\theta}(\mathbf X_t,\mathbf A_y;\mathbf E).
    \label{eq:base_forecast}
\end{equation}
Here, $\mathbf E$ contains the backbone's existing node-specific parameters.
STAEFormer uses an adaptive spatio-temporal embedding block for each
sensor~\cite{liu2023spatio}. STID uses spatial embeddings~\cite{shao2022spatial},
and Graph WaveNet uses two node embedding tables~\cite{wu2019graph}.
We write $\mathbf e_n$ for all identity components of sensor $n$.
Shared layers, output heads, shared temporal parameters, and normalization
statistics remain fixed. Identity fitting keeps the backbone in evaluation
mode. The input $\mathbf A_y$ is supplied when used by the backbone.

\paragraph{Identity refresh.}
Let $\mathcal R$ contain recent windows with released targets and
$\mathcal S$ the sensors being fitted.
We reserve the final chronological quarter of $\mathcal R$ for early
stopping and optimize on the earlier windows, $\mathcal R_{\mathrm{fit}}$:
\begin{equation}
    \begin{aligned}
    \mathcal L_{\mathrm{id}}(\mathbf E;\mathcal R,\mathcal S)
    &=\frac{1}{Z}
      \sum_{i\in\mathcal R_{\mathrm{fit}}}
      \sum_{n\in\mathcal S}\sum_{h=1}^{H}m_{i,n,h}\\
    &\qquad{}\cdot
      \left|\widehat y^{\mathrm{base}}_{i,n,h}(\mathbf E)-y_{i,n,h}\right|.
    \end{aligned}
    \label{eq:identity_refresh}
\end{equation}
The mask $m_{i,n,h}$ equals one for finite, nonzero targets, and
$Z$ counts these entries over the fitted windows and sensors.
When all active sensors have recent history, $\mathcal S=\mathcal V_y$:
matched checkpoint identities initialize $\mathcal K$, and the
componentwise mean checkpoint identity initializes $\mathcal U$.
For arrival preparation, we first fit $\mathcal K$ using inputs and any
adjacency restricted to that sensor set, then initialize $\mathcal U$
as described in Section~\ref{sec:arrival_refit}.

For Graph WaveNet, the maintained tables also define its adaptive adjacency:
\begin{equation}
    \mathbf A_{\mathrm{adp}}
    =\operatorname{softmax}_{\mathrm{row}}
    \left(\operatorname{ReLU}(\mathbf E_1\mathbf E_2^\top)\right).
    \label{eq:identity_adjacency}
\end{equation}
Refreshing $\mathbf E_1$ and $\mathbf E_2$ changes the learned adjacency.
The static adjacency is retained alongside it, with shared propagation
weights fixed.

\subsection{Causal Delayed-Feedback Pool}
\label{sec:causal_pool}
A forecast issued at $i$ covers $i+1{:}i+H$. Its record enters a pending
queue $\mathcal Q$ after prediction. Before predicting at $t$, eligible
records move to the released pool:
\begin{equation}
    \mathcal P_t=\mathcal P_{t-1}\cup
    \left\{q_i\in\mathcal Q\,\middle|\,i+H\leq t\right\}.
    \label{eq:causal_pool}
\end{equation}
The pool starts with eligible history available before the stream.
Each record stores its input, the base forecast computed at issue time,
sensor IDs, and origin index. Its target is attached upon release.
Released records leave $\mathcal Q$.
Identity fitting selects recent released windows. Ordinary RLS updates
consume each new eligible record once, in time order, using its cached
base forecast. Feedback awaiting an RLS update is tracked separately from
the history retained for fitting.

\subsection{New-Sensor Initialization and Joint Refitting}
\label{sec:arrival_refit}

\paragraph{Borrowing an identity.}
For $u\in\mathcal U$ and $|\mathcal K|>0$, we compare its short
post-arrival history with
contemporaneous observations from $v\in\mathcal K$.
Let $\mathcal O_{uv}$ contain the times available before initialization
when both sensors have finite readings. We use
\begin{equation}
    \begin{aligned}
    d(u,v)&=\frac{1}{|\mathcal O_{uv}|}
       \sum_{r\in\mathcal O_{uv}}(s_{r,u}-s_{r,v})^2,\\
    \mathbf e_u^{(0)}
       &=\frac{1}{k}\sum_{v\in\mathcal N_k(u)}\mathbf e_v,
    \end{aligned}
    \label{eq:identity_borrow}
\end{equation}
We set $k=\min(5,|\mathcal K|)$ and use the nearest eligible donors
as $\mathcal N_k(u)$. Averaging covers all identity components.
If fewer than $k$ donors meet the observation requirements, we draw the
entire donor set from $\mathcal K$ with a fixed random seed.
Borrowed identities remain fixed until the joint refit.
Appendix~\ref{app:evost_implementation} gives the eligibility rules.

\paragraph{Refitting the expanded network.}
At the scheduled time $t_\star$, we collect the latest $W$ released
windows, including eligible prestream history when needed.
Starting from the current identity table, we optimize
\cref{eq:identity_refresh} over all active sensors,
$\mathcal S=\mathcal K\cup\mathcal U$.
We then recompute base forecasts on those windows with the updated
identities and rebuild every calibrator's coefficients and inverse
covariance state by ridge regression.
The selected ridge penalty, forgetting factor, and output scale are retained.
Pending records keep their original cached forecasts.
After rebuilding, recursive updates resume with newly released feedback.
The refitting history is not repeatedly consumed by RLS.

\subsection{Per-Sensor Recursive Calibration}
\label{sec:recursive_calibration}
For window $t$ and sensor $n$, let
$\mathbf b_{t,n}=\widehat{\mathbf y}^{\mathrm{base}}_{t,n}$.
We stack the scaled base forecast, checkpoint-normalized history
$\bar{\mathbf x}_{t,n}$, four temporal summaries, and an intercept
into a column vector:
\begin{equation}
    \mathbf z_{t,n}
    =\operatorname{col}\!\left(
      \mathbf b_{t,n}/s,\bar{\mathbf x}_{t,n},
      \operatorname{Stat}(\bar{\mathbf x}_{t,n}),1\right).
    \label{eq:rls_features}
\end{equation}
The summaries are the last value, mean, population standard deviation,
and least-squares slope of the history.
The global output scale $s$ is fixed from valid targets in the initial
training tail. Appendix~\ref{app:evost_implementation} specifies its scope
in the arrival evaluation.
With $D=H+L+5$, a separate matrix
$\boldsymbol\Theta_n\in\mathbb R^{D\times H}$ corrects each sensor:
\begin{equation}
    \widehat{\mathbf y}_{t,n}
    =\mathbf b_{t,n}+s\boldsymbol\Theta_n^\top\mathbf z_{t,n}.
    \label{eq:recursive_prediction}
\end{equation}
Let $\mathcal J_n$ contain fitting records whose $H$ targets are all finite
and nonzero and which satisfy the sensor's arrival restriction.
With $\mathbf r_{i,n}=(\mathbf y_{i,n}-\mathbf b_{i,n})/s$
and ridge penalty $\rho>0$, we initialize
\begin{equation}
    \begin{aligned}
    \mathbf P_n^{(0)}
      &=\left(\rho\mathbf I_D+
        \sum_{i\in\mathcal J_n}\mathbf z_{i,n}\mathbf z_{i,n}^\top
        \right)^{-1},\\
    \boldsymbol\Theta_n^{(0)}
      &=\mathbf P_n^{(0)}
        \sum_{i\in\mathcal J_n}\mathbf z_{i,n}\mathbf r_{i,n}^\top.
    \end{aligned}
    \label{eq:ridge_initialization}
\end{equation}
Calibration uses the full eligible fitting history, including the quarter
reserved for identity early stopping.
For a newly eligible record, RLS updates
\begin{align}
    \mathbf k_n
      &=\frac{\mathbf P_n\mathbf z_{i,n}}
        {\lambda+\mathbf z_{i,n}^\top\mathbf P_n\mathbf z_{i,n}},
        \nonumber\\
    \boldsymbol\epsilon_{i,n}
      &=\mathbf r_{i,n}-\boldsymbol\Theta_n^\top\mathbf z_{i,n},
        \nonumber\\
    \boldsymbol\Theta_n^+
      &=\boldsymbol\Theta_n+\mathbf k_n\boldsymbol\epsilon_{i,n}^\top,
        \nonumber\\
    \mathbf P_n^+
      &=\lambda^{-1}
        \left(\mathbf P_n-\mathbf k_n\mathbf z_{i,n}^\top\mathbf P_n\right),
    \label{eq:recursive_update}
\end{align}
where $\lambda\in(0,1]$ and the right-hand sides use the state before
the update. The same complete-target validity rule applies to initialization
and recursion. These online updates change only the calibration state.
Identities remain fixed between fitting stages.

\subsection{Unified Online Procedure}
\label{sec:online_procedure}
\Cref{alg:evost}\extref{}{ (Appendix~\ref{app:evost_implementation})} summarizes the feedback order after identity preparation,
calibration initialization, and validation-based hyperparameter selection.
The arrival refit coincides with an RLS update,
$t_\star\bmod\delta=0$, so all feedback released by then is processed
before refitting. Test predictions follow the complete-horizon release rule.

\newcommand{\EvoSTAlgorithm}{%
\begin{algorithm}[t]
\caption{EvoST with delayed feedback}
\label{alg:evost}
\footnotesize
\begin{algorithmic}[1]
\REQUIRE Frozen $\boldsymbol\theta$; prepared $\mathbf E$ and
$\{(\boldsymbol\Theta_n,\mathbf P_n)\}$; eligible history $\mathcal H_{\mathrm{pre}}$;
fixed $\rho,\lambda,s$; delay $H$; cadence $\delta$; optional
$t_\star$ with $t_\star\bmod\delta=0$; refit length $W$
\ENSURE Forecasts $\{\widehat{\mathbf Y}_t\}$ using released feedback
\STATE $\mathcal Q\gets\emptyset$; $\mathcal P\gets\mathcal H_{\mathrm{pre}}$;
$\mathcal I\gets\emptyset$ \COMMENT{feedback awaiting RLS}
\FOR{each chronological window $t=0,1,\ldots$}
    \STATE Move records with $i+H\leq t$ from $\mathcal Q$ to $\mathcal P$ and $\mathcal I$, attaching their targets
    \IF{$t\bmod\delta=0$}
        \STATE Update RLS on valid records in $\mathcal I$ in time order using cached base forecasts and \cref{eq:recursive_update}
        \STATE $\mathcal I\gets\emptyset$
    \ENDIF
    \IF{refitting is enabled and $t=t_\star$}
        \STATE $\mathcal R\gets\operatorname{Tail}_{W}(\mathcal P)$
        \STATE Refit all active identities from the current $\mathbf E$ using \cref{eq:identity_refresh}
        \STATE Recompute base forecasts, features, and residual targets on $\mathcal R$
        \STATE Rebuild all $(\boldsymbol\Theta_n,\mathbf P_n)$ using \cref{eq:ridge_initialization}, retaining $\rho,\lambda,s$
    \ENDIF
    \STATE $\widehat{\mathbf Y}^{\mathrm{base}}_t\gets f_{\boldsymbol\theta}(\mathbf X_t,\mathbf A_y;\mathbf E)$
    \STATE Correct each sensor's forecast using \cref{eq:recursive_prediction}
    \STATE Emit $\widehat{\mathbf Y}_t$; enqueue the record and its base forecast in $\mathcal Q$ with release time $t+H$
\ENDFOR
\end{algorithmic}
\end{algorithm}}
\ifextended
\EvoSTAlgorithm
\fi

\section{Experiments}
\label{exp}

\newcommand{\SensitivityFigure}{%
\begin{figure*}[t]
    \centering
    \includegraphics[width=0.9686\textwidth]{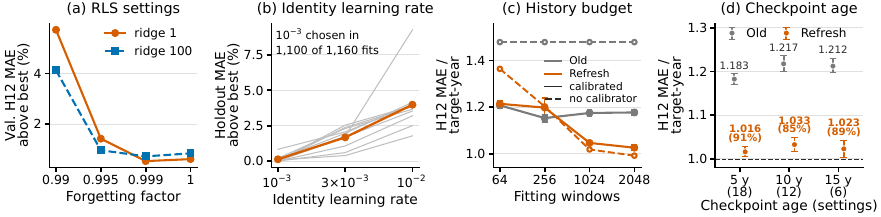}
    \caption{Sensitivity. (a, b) Validation (RLS) and holdout (learning rate) MAE above each run's best (1,165 runs, 1,160 fits). (c) History budget, calibrated (solid) or uncalibrated (dashed). (d) Checkpoint ages on different target years.}
    \label{fig:sensitivity}
    \Description{Four panels show the validation error of the RLS settings against the forgetting factor, the holdout error against the identity learning rate, error ratios against the number of fitting windows, and old and refreshed checkpoints at three ages.}
\end{figure*}}

\newcommand{\AblationFigure}{%
\ifextended
\begin{figure*}[t]
    \centering
    \includegraphics[width=\textwidth]{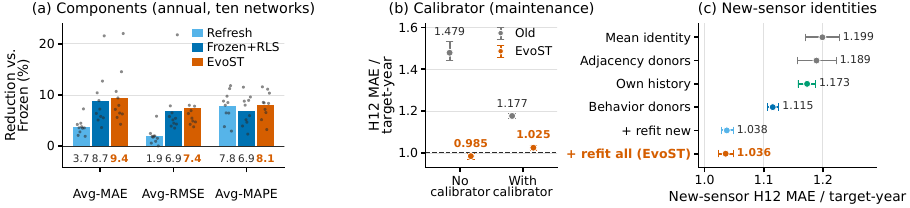}
\else
\begin{figure}[t]
    \centering
    \includegraphics[width=\linewidth]{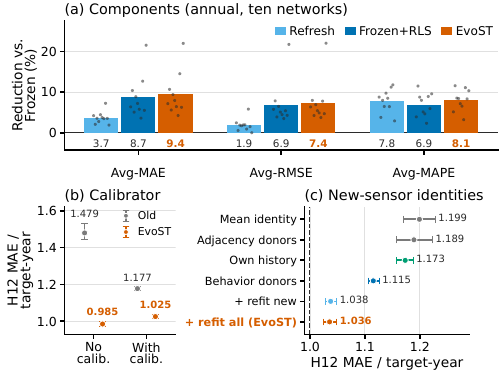}
\fi
    \caption{Ablations. Panels show (a) reductions from frozen per-year checkpoints on ten networks (bars: equal-weight means, dots: networks), (b) Old and EvoST, all calibrated or uncalibrated, and (c) calibrated new-sensor results after the arrival refit.}
    \label{fig:ablation}
    \Description{Three panels show reductions relative to the frozen checkpoints for refresh only, calibration only, and both, the old checkpoint and EvoST with and without calibration, and six ways to initialize or refit new-sensor identities.}
\ifextended
\end{figure*}
\else
\end{figure}
\fi}

\subsection{Experimental Settings}
PeMS experiments use the first 31 days of each selected year.
Table~\ref{tab:evo_scale} summarizes the full data archive.
Each year uses a chronological 60\%/20\%/20\% train/validation/test split.
PEMS windows contain 12 five-minute steps and TFNSW windows contain 12 hourly
steps. In the annual benchmark, conventional forecasters share graph
snapshots, splits, raw input and target windows, and metric definitions.
Conventional baseline training uses the benchmark's cached inputs, which
are standardized within each split using that split's statistics, following
TrafficStream~\cite{ijcai2021p0498}. The EvoST pipeline rebuilds
validation and test inputs from raw windows and standardizes them with the
statistics of the model's training split. Residual non-finite inputs are set
to zero after normalization, and invalid targets are masked. Foundation
models receive raw-scale contexts and retain their model-specific input
normalization.
\paragraph{Checkpoint maintenance.}
These experiments reuse the per-year STAEFormer checkpoints of the annual
benchmark below (checkpoint seeds 42--44) on later target years, with the
statistics of each checkpoint's training split. The main panel applies
five-year-old checkpoints to PEMS03, PEMS06, PEMS08, PEMS10, PEMS11, and
PEMS12 in 2020 and 2025, giving twelve region-year settings. Parameter
updates, including the prompt and scratch controls, draw from the latest
2,048 training windows of the target year, with the final 512 reserved for
early stopping and learning-rate selection. \extref{Both the fitting and holdout
subsets are sampled at a stride of two (four above 1,600 sensors). Each
variant that updates parameters in the main comparison selects its learning
rate on this holdout. No validation or test value enters these initial fits. The
scratch control uses target-year training statistics. All variants and the
reference then use the same per-sensor RLS procedure, with labels released
after 12 steps and updates every 64 windows. Its penalty and forgetting
factor are selected separately for each variant on the first 1,024
validation windows. After reinitialization from the training tail, RLS
processes validation and test chronologically. Only test errors are
reported. The reference is the target-year STAEFormer checkpoint with this
calibration. We report MAE over the one-hour horizon, accumulated over steps
1--12 (H12). Errors are averaged over evaluation seeds 51--53 within each
setting, divided by the reference, and combined by the geometric mean across
settings. The 95\% intervals use 10,000 bootstrap draws over districts, keeping
both years of a district together. Extensions to other checkpoint ages,
backbones, sensor arrivals, and label budgets state their panels and any
fixed learning rates with their results. Appendix~\ref{app:evost_implementation}
provides implementation details.}{Fitting and holdout windows are sampled at a stride of two (four above 1,600 sensors), each updating variant selects its learning rate on the holdout, and no validation or test value enters these fits. The scratch control uses target-year training statistics. All variants and the reference, the target-year checkpoint, then use the per-sensor RLS procedure (\Cref{sec:recursive_calibration}) with settings selected per variant on the first 1,024 validation windows. RLS is then reinitialized from the training tail, validation and test are replayed chronologically, and only test errors are reported. We report MAE accumulated over steps 1--12 (H12), averaged over evaluation seeds 51--53, divided by the reference, and combined by the geometric mean over settings, with 95\% intervals from 10,000 bootstrap draws over districts (Appendix~\ref{app:evost_implementation}).}
\paragraph{Annual benchmark.}
For each year, EvoST starts from the benchmark's STAEFormer checkpoint of
that year, so every sensor of the year already has an identity. \extref{It refreshes
the identity table on up to the latest 2,048 training windows, using the
final 512 window starts for early stopping and for selecting the learning
rate from $\{10^{-3},3\times10^{-3},10^{-2}\}$, and fits each sensor's
calibrator on the training tail. Calibration settings are selected on up to
the first 1,024 validation windows. Validation labels are used in this
selection, and the resulting RLS states are discarded. New calibrators fitted
on the training tail then run over the test portion only, using each target once
it is released 12 steps later and updating every 64 windows. Static
forecasters do not adapt to test labels. Appendix~\ref{app:annual} gives the
remaining settings and matched controls.}{It refreshes the identities on up to the latest 2,048 training windows (the final 512 for early stopping and for choosing the learning rate from $\{10^{-3},3\times10^{-3},10^{-2}\}$) and fits each sensor's calibrator on the training tail. Calibration settings are selected on up to the first 1,024 validation windows. New calibrators fitted on the training tail then run over the test portion only, using each target once it is released 12 steps later and updating every 64 windows (Appendix~\ref{app:annual}). Static forecasters do not adapt to test labels.} We report cumulative mean absolute
error (MAE), root mean square error (RMSE), and mean absolute percentage
error (MAPE) at horizons 3, 6, and 12. Avg averages the cumulative scores
from horizons 1 to 12, with Avg-MAE as the primary metric. Results use five
seeds unless noted otherwise.

\subsection{Baselines}
\paragraph{Maintenance controls.}
On five-year-old STAEFormer checkpoints, we compare output-head tuning
(Head), shared-weight tuning (Shared), full-parameter fine-tuning (Full FT),
an EAC-style low-rank input prompt~\cite{chen2025expand} (Prompt), node-identity
refresh (EvoST), and joint identity and prompt tuning (EvoST+Prompt).
Old keeps the forecaster parameters fixed, while Scratch trains a randomly
initialized model on the same target-year fitting windows. \extref{All variants and
the target-year checkpoint reference use the per-sensor RLS procedure in
Section~\ref{sec:recursive_calibration}.}{}
\extref{On the annual benchmark, we compare nine static forecasters (DCRNN, ASTGNN, TGCN, GWN, STID, STNorm,
iTrans, DLinear, and STAEFormer), four naive schemes (Pretrain, Retrain, Online-NN, and Online-AN),
and four evolving-graph methods (TrafficStream, PECPM, STKEC, and EAC). We also compare STRAP,
ST-TTC, and zero-shot (ZS) or yearly fine-tuned (FT) versions of Chronos-2,
TimesFM~2.5, and Moirai-2.0.}{On the annual benchmark, we compare the nine static forecasters, four naive schemes, four evolving-graph methods, STRAP, ST-TTC, and zero-shot (ZS) or yearly fine-tuned (FT) Chronos-2, TimesFM~2.5, and Moirai-2.0 listed in \Cref{tab:main_by_method_part1}.}
EvoST starts from stored per-year STAEFormer checkpoints of the benchmark.
\extref{The Online-AN and STAEFormer rows report the benchmark's test errors recorded
at the end of training on split-specific cached inputs. The STAEFormer row includes
later single-year retraining for PEMS04/2002 and TFNSW/2006. The EvoST row
rebuilds inputs with training-split statistics, and its selected STAEFormer
checkpoints differ in some cases from those behind the STAEFormer row.}{The Online-AN and STAEFormer rows report the benchmark's test errors at the end of training on split-specific cached inputs. The STAEFormer row includes later single-year retraining for PEMS04/2002 and TFNSW/2006, and its checkpoints differ in some cases from those EvoST starts from.}

\subsection{Which Parameters Need Maintenance?}
\label{sec:which_params}

\providecommand{\best}[1]{\cellcolor{yellow!35}\textbf{\textcolor{red}{#1}}}
\providecommand{\second}[1]{\cellcolor{black!10}\underline{#1}}
\begin{table*}[!h]
\centering
\caption{Avg results on PEMS03, PEMS04, and TFNSW: mean $\pm$ SD (sample SD over five seeds for EvoST, none for the deterministic zero-shot runs). MAPE is in percent. Bold and underline mark the best and second-best non-foundation methods. Foundation models are not ranked. Chronos-2 uses multivariate graph-grouped inference, and its fine-tuned variant uses per-year LoRA.}
\label{tab:main_by_method_part1}
\setlength{\tabcolsep}{2.4pt}
\footnotesize
\renewcommand{\arraystretch}{0.92}
\resizebox{\textwidth}{!}{%
\begin{tabular}{ll|ccc|ccc|ccc}
\toprule
\multirow{2}{*}{\textbf{Category}} & \multirow{2}{*}{\textbf{Baseline}} & \multicolumn{3}{c|}{\textbf{PEMS03}} & \multicolumn{3}{c|}{\textbf{PEMS04}} & \multicolumn{3}{c}{\textbf{TFNSW}} \\
\cmidrule(lr){3-5} \cmidrule(lr){6-8} \cmidrule(lr){9-11}
 & & \textbf{MAE} & \textbf{RMSE} & \textbf{MAPE (\%)} & \textbf{MAE} & \textbf{RMSE} & \textbf{MAPE (\%)} & \textbf{MAE} & \textbf{RMSE} & \textbf{MAPE (\%)} \\
\midrule
\multirow{4}{*}{\textbf{\shortstack[l]{Static\\STGNN\\Backbones}}} & DCRNN & 14.99$_{\pm0.18}$ & 24.90$_{\pm0.19}$ & 28.69$_{\pm0.86}$ & 22.22$_{\pm0.27}$ & 34.58$_{\pm0.32}$ & 20.53$_{\pm0.40}$ & 179.87$_{\pm14.50}$ & 303.16$_{\pm19.20}$ & 211.72$_{\pm42.72}$ \\
 & ASTGNN & 15.15$_{\pm0.10}$ & 25.15$_{\pm0.17}$ & 26.56$_{\pm0.43}$ & 22.45$_{\pm0.09}$ & 34.96$_{\pm0.12}$ & 19.29$_{\pm0.42}$ & 140.15$_{\pm0.82}$ & 247.22$_{\pm1.02}$ & 125.28$_{\pm3.02}$ \\
 & TGCN & 15.88$_{\pm0.10}$ & 30.12$_{\pm0.40}$ & 28.95$_{\pm0.89}$ & 23.60$_{\pm0.37}$ & 38.82$_{\pm0.55}$ & 20.15$_{\pm0.28}$ & 208.33$_{\pm1.48}$ & 390.75$_{\pm3.25}$ & 177.95$_{\pm3.89}$ \\
 & GWN & 15.33$_{\pm0.26}$ & 25.15$_{\pm0.39}$ & 28.01$_{\pm0.71}$ & 23.41$_{\pm0.27}$ & 35.69$_{\pm0.22}$ & 19.94$_{\pm0.55}$ & 170.89$_{\pm3.29}$ & 313.61$_{\pm5.86}$ & 138.49$_{\pm10.57}$ \\
\midrule
\multirow{4}{*}{\textbf{\shortstack[l]{Na\"ive \\Schemes}}} & Pretrain & 31.37$_{\pm1.11}$ & 41.64$_{\pm0.99}$ & 50.04$_{\pm2.80}$ & 153.33$_{\pm6.62}$ & 186.37$_{\pm7.58}$ & 82.25$_{\pm2.99}$ & 262.63$_{\pm10.87}$ & 383.38$_{\pm12.29}$ & 430.83$_{\pm35.95}$ \\
 & Retrain & 14.16$_{\pm0.15}$ & 23.46$_{\pm0.14}$ & 26.29$_{\pm0.58}$ & 21.64$_{\pm0.20}$ & 33.58$_{\pm0.22}$ & 19.15$_{\pm0.30}$ & 138.04$_{\pm0.62}$ & 247.58$_{\pm1.00}$ & 104.00$_{\pm3.59}$ \\
 & Online-NN & 17.47$_{\pm0.16}$ & 28.29$_{\pm0.08}$ & 29.62$_{\pm1.04}$ & 38.46$_{\pm0.27}$ & 51.85$_{\pm0.46}$ & 26.93$_{\pm0.71}$ & 174.36$_{\pm3.31}$ & 318.17$_{\pm5.87}$ & 129.82$_{\pm8.24}$ \\
 & Online-AN & 13.80$_{\pm0.05}$ & 22.89$_{\pm0.11}$ & 24.96$_{\pm0.21}$ & 21.01$_{\pm0.26}$ & 32.73$_{\pm0.31}$ & 17.82$_{\pm0.31}$ & 130.61$_{\pm0.96}$ & 235.35$_{\pm1.95}$ & 99.93$_{\pm3.54}$ \\
\midrule
\multirow{4}{*}{\textbf{\shortstack[l]{Evolving\\Graph}}} & TrafficStream & 14.32$_{\pm0.20}$ & 23.60$_{\pm0.28}$ & 27.63$_{\pm1.67}$ & 21.43$_{\pm0.53}$ & 33.35$_{\pm0.60}$ & 18.83$_{\pm0.59}$ & 145.42$_{\pm2.41}$ & 259.40$_{\pm4.67}$ & 106.92$_{\pm8.01}$ \\
 & PECPM & 14.90$_{\pm0.81}$ & 24.32$_{\pm0.98}$ & 27.78$_{\pm2.30}$ & 25.16$_{\pm3.18}$ & 39.17$_{\pm4.12}$ & 22.29$_{\pm3.69}$ & 174.19$_{\pm22.07}$ & 318.61$_{\pm45.67}$ & 106.24$_{\pm4.44}$ \\
 & STKEC & 14.47$_{\pm0.22}$ & 23.89$_{\pm0.34}$ & 27.58$_{\pm0.64}$ & 22.78$_{\pm1.14}$ & 34.78$_{\pm1.18}$ & 21.54$_{\pm1.53}$ & 147.86$_{\pm1.05}$ & 259.82$_{\pm2.21}$ & 148.01$_{\pm7.69}$ \\
 & EAC & 15.91$_{\pm0.48}$ & 26.40$_{\pm0.88}$ & 30.12$_{\pm0.81}$ & 34.02$_{\pm9.03}$ & 48.20$_{\pm9.87}$ & 39.19$_{\pm17.54}$ & 137.90$_{\pm1.42}$ & 245.35$_{\pm1.93}$ & 119.37$_{\pm14.85}$ \\
\midrule
\multirow{2}{*}{\textbf{\shortstack[l]{Retrieval\\TTC}}} & STRAP & 14.15$_{\pm0.13}$ & 23.37$_{\pm0.22}$ & 26.12$_{\pm0.31}$ & 21.93$_{\pm0.67}$ & 33.78$_{\pm0.64}$ & 19.49$_{\pm0.57}$ & 135.57$_{\pm1.80}$ & 243.47$_{\pm4.15}$ & 104.21$_{\pm3.09}$ \\
 & ST-TTC & 13.98$_{\pm0.14}$ & 23.18$_{\pm0.13}$ & 26.15$_{\pm0.57}$ & 21.33$_{\pm0.20}$ & 33.21$_{\pm0.22}$ & 19.04$_{\pm0.29}$ & 144.41$_{\pm0.72}$ & 257.96$_{\pm1.08}$ & 106.33$_{\pm3.01}$ \\
\midrule
\multirow{5}{*}{\textbf{\shortstack[l]{Static\\Forecasting\\Backbones}}} & STID & 15.14$_{\pm0.14}$ & 27.10$_{\pm0.38}$ & 26.44$_{\pm0.62}$ & 22.43$_{\pm0.18}$ & 35.63$_{\pm0.31}$ & 18.51$_{\pm0.26}$ & 153.63$_{\pm6.20}$ & 307.19$_{\pm10.11}$ & 112.25$_{\pm11.68}$ \\
 & STNorm & 18.20$_{\pm0.47}$ & 28.58$_{\pm0.65}$ & 52.94$_{\pm2.57}$ & 23.99$_{\pm0.62}$ & 36.65$_{\pm0.80}$ & 27.98$_{\pm1.51}$ & 148.78$_{\pm2.33}$ & 275.25$_{\pm3.16}$ & 182.00$_{\pm6.10}$ \\
 & iTrans & 14.21$_{\pm0.13}$ & 23.42$_{\pm0.14}$ & 26.79$_{\pm0.71}$ & 21.22$_{\pm0.11}$ & 33.04$_{\pm0.17}$ & 19.36$_{\pm0.25}$ & 144.04$_{\pm0.82}$ & 257.27$_{\pm2.16}$ & 113.97$_{\pm6.78}$ \\
 & DLinear & 15.73$_{\pm0.06}$ & 26.23$_{\pm0.04}$ & 28.55$_{\pm0.98}$ & 23.58$_{\pm0.18}$ & 36.54$_{\pm0.22}$ & 19.60$_{\pm0.33}$ & 182.20$_{\pm1.03}$ & 318.13$_{\pm1.10}$ & 187.17$_{\pm5.55}$ \\
 & STAEFormer & \second{12.26$_{\pm0.11}$} & \second{20.58$_{\pm0.12}$} & \second{23.06$_{\pm0.64}$} & \second{19.38$_{\pm0.94}$} & \second{30.64$_{\pm1.20}$} & \second{16.97$_{\pm0.95}$} & \second{96.12$_{\pm1.67}$} & \second{190.33$_{\pm2.83}$} & \second{66.43$_{\pm2.35}$} \\
\midrule
\multirow{6}{*}{\textbf{\shortstack[l]{Times-series\\Forecasting\\Foundation\\Model}}} & Chronos-2 (ZS) & 15.27$_{\pm0.00}$ & 26.98$_{\pm0.00}$ & 25.83$_{\pm0.00}$ & 21.58$_{\pm0.00}$ & 35.22$_{\pm0.00}$ & 17.62$_{\pm0.00}$ & 227.74$_{\pm0.00}$ & 440.18$_{\pm0.00}$ & 120.67$_{\pm0.00}$ \\
 & Chronos-2 (FT) & 13.31$_{\pm0.01}$ & 23.66$_{\pm0.03}$ & 22.80$_{\pm0.03}$ & 18.94$_{\pm0.04}$ & 31.89$_{\pm0.12}$ & 15.25$_{\pm0.01}$ & 160.20$_{\pm0.68}$ & 323.47$_{\pm2.06}$ & 78.26$_{\pm0.28}$ \\
 & TimesFM2.5 (ZS) & 15.91$_{\pm0.00}$ & 27.86$_{\pm0.00}$ & 26.71$_{\pm0.00}$ & 23.49$_{\pm0.00}$ & 37.69$_{\pm0.00}$ & 19.32$_{\pm0.00}$ & 266.77$_{\pm0.00}$ & 487.06$_{\pm0.00}$ & 160.45$_{\pm0.00}$ \\
 & TimesFM2.5 (FT) & 14.16$_{\pm0.01}$ & 25.36$_{\pm0.04}$ & 23.79$_{\pm0.01}$ & 20.92$_{\pm0.02}$ & 35.20$_{\pm0.02}$ & 16.65$_{\pm0.02}$ & 215.22$_{\pm0.33}$ & 420.65$_{\pm0.52}$ & 124.25$_{\pm0.83}$ \\
 & Moirai-2.0 (ZS) & 16.19$_{\pm0.00}$ & 28.00$_{\pm0.00}$ & 26.15$_{\pm0.00}$ & 24.52$_{\pm0.00}$ & 38.87$_{\pm0.00}$ & 19.11$_{\pm0.00}$ & 294.05$_{\pm0.00}$ & 545.09$_{\pm0.00}$ & 137.97$_{\pm0.00}$ \\
 & Moirai-2.0 (FT) & 13.95$_{\pm0.01}$ & 24.51$_{\pm0.03}$ & 24.18$_{\pm0.01}$ & 20.68$_{\pm0.03}$ & 34.42$_{\pm0.09}$ & 16.78$_{\pm0.02}$ & 142.36$_{\pm0.28}$ & 303.78$_{\pm0.69}$ & 62.69$_{\pm0.22}$ \\
\midrule
\textbf{Ours} & EvoST & \best{10.92$_{\pm0.05}$} & \best{18.47$_{\pm0.07}$} & \best{22.52$_{\pm0.21}$} & \best{16.58$_{\pm0.21}$} & \best{27.14$_{\pm0.26}$} & \best{15.52$_{\pm0.23}$} & \best{65.16$_{\pm0.34}$} & \best{137.77$_{\pm0.59}$} & \best{52.89$_{\pm0.29}$} \\
\bottomrule
\end{tabular}%
}
\end{table*}

\begin{figure}[!b]
    \centering
    \includegraphics[width=\linewidth]{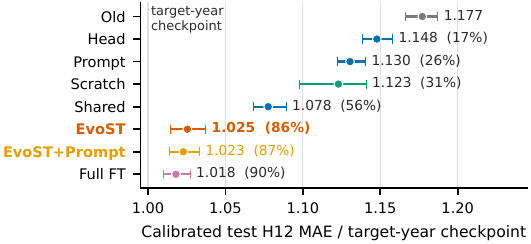}
    \caption{Parameter maintenance of five-year-old STAEFormer checkpoints (Scratch: trained anew). Calibrated test H12 MAE ratios to the calibrated target-year checkpoint (dashed) are geometric means over twelve settings after averaging three seeds, with 95\% district-bootstrap intervals. Percentages give the share of the Old gap recovered.}
    \label{fig:maintenance}
    \Description{Dots show eight maintenance variants with interval bars. The five-year-old checkpoint has ratio 1.177, output-head tuning 1.148, the prompt control 1.130, Scratch 1.123, shared-weight tuning 1.078, identity refresh 1.025, identity refresh with prompt 1.023, and full fine-tuning 1.018.}
\end{figure}

\extref{\Cref{fig:maintenance} shows the maintenance controls on the main panel.
The five-year-old checkpoint (Old) is 17.7\% worse than the target-year
checkpoint. Refreshing only the node identities (EvoST) lowers the ratio to
1.025 and recovers 86\% of this gap. Split by sensor, identity refresh
recovers 85\% of the gap on the 81\% of sensors the old checkpoint knows
and 88\% on the rest, which start from the mean identity, and it beats Old
and Shared in all twelve settings for both groups (\Cref{tab:known_unseen}).
Full fine-tuning recovers slightly more
(1.018): identity refresh is 0.7\% worse than full fine-tuning (95\% interval
0.3--1.3\%). Tuning only the shared spatio-temporal weights (1.078) or the
output head (1.148) recovers much less, as do a rank-6 EAC-style input prompt
(1.130) and a model trained from scratch on the same target-year windows
(1.123). Refitting the prompt jointly with the identities reduces MAE by a
further 0.25\% relative to EvoST. On the H100 runs, the median ratio of epoch
time for identity refresh to Full FT is 0.82, because fitting still backpropagates
through the frozen backbone. Identity refresh changes only per-node parameters
and leaves the shared model unchanged. These comparisons use the specified
fitting budgets and learning-rate grids.
Appendix~\ref{app:evost_implementation} reports the settings and selection
counts.}{On the main panel (\Cref{fig:maintenance}), the five-year-old checkpoint (Old) is 17.7\% worse than the target-year checkpoint. Refreshing only the node identities (EvoST) recovers 86\% of this gap. Split by sensor, it recovers 85\% of the gap on the 81\% of sensors the old checkpoint knows and 88\% on the rest, beating Old and Shared in every setting for both groups. Full fine-tuning recovers slightly more: identity refresh is 0.7\% worse (95\% interval 0.3--1.3\%). Tuning only the shared weights or the output head recovers much less, as do a rank-6 EAC-style input prompt and a model trained from scratch. Identity refresh leaves the shared model unchanged (Appendix~\ref{app:evost_implementation}).} On the main STAEFormer panel, median parameter counts are 1.41M in the identity table, 0.60M shared, and 2.01M total. The rank-6 input prompt has a median of 8,856 trainable values. These controls compare update strategies with different parameter budgets.

\subsection{Other Backbones and Checkpoint Ages}
\label{sec:age_backbones}

\begin{table}[t]
\centering
\caption{Five-year reuse on sixteen region-year settings (eight districts, 2020 and 2025). Calibrated test H12 MAE ratios to each backbone's calibrated target-year checkpoint are geometric means after averaging three seeds per setting. STAEFormer and STID use a fixed refresh learning rate of $3\times10^{-3}$. Recovery is the fraction of the old-to-reference error gap closed.}
\label{tab:backbones}
\begin{tabular}{lrrrr}
\toprule
Backbone & Settings & Old & Refresh & Recovery \\
\midrule
STAEFormer & 16 & 1.180 & 1.033 & 82\% \\
STID & 16 & 1.045 & 1.015 & 67\% \\
Graph WaveNet & 16 & 1.091 & 1.011 & 88\% \\
\bottomrule
\end{tabular}
\end{table}

The finding is not specific to STAEFormer (\Cref{tab:backbones}). \extref{On sixteen settings from
eight districts, five-year-old STID and Graph WaveNet checkpoints are 4.5\%
and 9.1\% worse than the target-year ones, against 18.0\% for STAEFormer.
Refreshing the node embeddings of STID and the node vectors that generate the
adaptive adjacency of Graph WaveNet recovers 67\% and 88\% of these gaps.
The benchmark's STAEFormer has no time-of-day or day-of-week inputs, so its
identities could absorb daily profiles. When STAEFormer is retrained per year with these
inputs, a five-year-old checkpoint on the main panel is still 14.0\% worse than
the target-year one (20.5\% for matched retraining without these inputs). Identity refresh recovers 90\% of this gap
(Appendix~\ref{app:evost_implementation}).}{On sixteen settings from eight districts, refreshing STID's node embeddings and the node vectors behind Graph WaveNet's adaptive adjacency recovers 67\% and 88\% of their smaller five-year gaps. The benchmark's STAEFormer has no time-of-day or day-of-week inputs, so its identities could absorb daily profiles. With matched yearly retraining on the main panel, the five-year MAE gap is 14.0\% with time inputs (20.5\% without). Identity refresh recovers 90\% of the former gap (Appendix~\ref{app:evost_implementation}).} STAEFormer checkpoints that are five, ten, and fifteen years old show similar recovery: refresh closes 91\%, 85\%, and 89\% of their gaps (\Cref{fig:sensitivity}(d)).

\subsection{Newly Added Sensors}
\label{sec:new_sensors}

\newcommand{\ArrivalFigure}[1]{%
\begin{figure}[t]
    \centering
    \includegraphics[width=#1\linewidth]{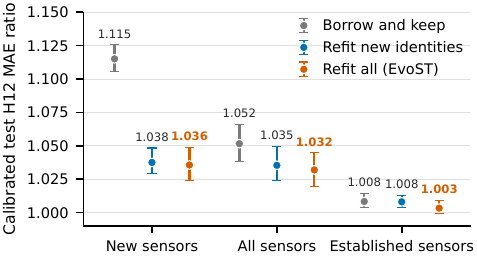}
    \caption{Sensor arrivals on twelve region-year settings (six districts, 2020 and 2025): calibrated test H12 MAE ratios, geometric means of three-seed averages, with 95\% district-bootstrap intervals. New and All use the target-year checkpoint as reference. Established uses the reference whose backbone runs on established sensors alone (Appendix~\ref{app:evost_implementation}).}
    \label{fig:arrival}
    \Description{Grouped dot plot for new, all, and established sensors. Borrowing and keeping the identities gives 1.115, 1.052, and 1.008. Refitting the new identities gives 1.038, 1.035, and 1.008. Refitting all identities, as EvoST does, gives 1.036, 1.032, and 1.003.}
\end{figure}}
\ifextended
\ArrivalFigure{1}
\fi

\extref{\Cref{fig:arrival} tracks}{We track} sensors that the five-year-old checkpoint has
\extref{never seen. Each arrives with 64 windows of history before the stream and
borrows the mean identity of its five behaviorally nearest established
sensors. At stream window 1,024, EvoST refits all identities and rebuilds
every calibrator (\Cref{sec:arrival_refit}). On the test portion, which lies
entirely after this refit, keeping the borrowed identities leaves new sensors
11.5\% worse than the target-year checkpoint. Refitting only the new
identities and rebuilding their calibrators lowers this to 3.8\% but leaves
established sensors 0.8\% worse than the calibrated reference whose backbone
runs on established sensors alone, with their identities refreshed before the
stream.
Refitting all identities, as EvoST does, gives 3.6\% and 0.3\%.
Rebuilding all calibrators in the new-only refit leaves established-sensor
MAE 0.9\% above its reference. Against this control, the all-identity refit
lowers it in 8 of 12 settings (geometric-mean change $-0.5\%$,
95\% interval $-1.1\%$ to $+0.1\%$).
The borrowed
start matters: kept unchanged, it gives 5.0\% lower MAE than independently
fitting new identities from the established-sensor mean on the same short
histories and 7.0\% lower MAE than the mean identity, whereas donors chosen
by road adjacency improve on the mean identity by only 0.8\%
(\Cref{fig:ablation}(c)).
Appendix~\ref{app:evost_implementation} describes how the all-identity refit
was specified.}{never seen. Each arrives with 64 windows of history and borrows the mean identity of its five behaviorally nearest sensors. At stream window 1,024, EvoST refits all identities and rebuilds every calibrator. On the test portion, after this refit, keeping the borrowed identities leaves new sensors 11.5\% worse than the target-year checkpoint. Refitting only the new identities and rebuilding their calibrators lowers this to 3.8\% but leaves established sensors 0.8\% worse than their calibrated reference, and refitting all identities gives 3.6\% and 0.3\%. Rebuilding all calibrators in the new-only refit leaves established-sensor MAE 0.9\% above its reference. Against this control, the all-identity refit lowers it in 8 of 12 settings (geometric-mean change $-0.5\%$, 95\% interval $-1.1\%$ to $+0.1\%$). Kept unchanged, borrowing outperforms the mean identity by 7.0\% (\Cref{fig:ablation}(c), Appendix~\ref{app:evost_implementation}).}

\subsection{Annual Benchmark}
\label{sec:annual_results}

\ifextended
\begin{figure*}[t]
    \centering
    \includegraphics[width=\textwidth]{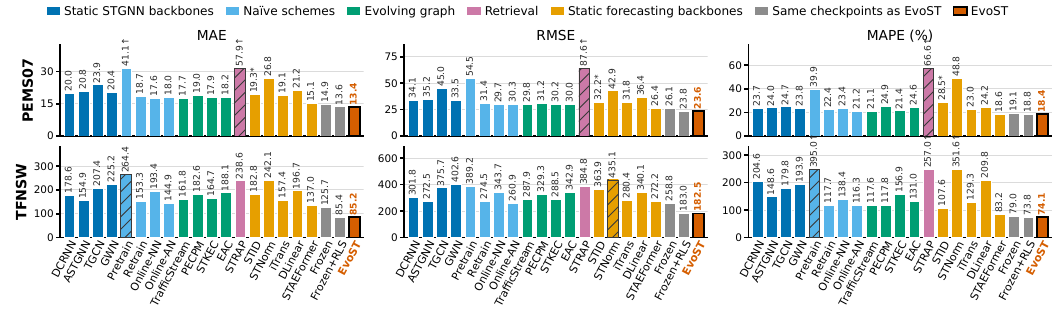}
    \caption{Sensors absent from the preceding year's graph (annual benchmark): H12 errors averaged over 19 arrival years, then five seeds. Frozen, Frozen+RLS, and EvoST use training-split input statistics. Other bars use split-specific statistics. ST-TTC is omitted: its archived new-sensor rows duplicate Retrain. *PEMS07 STID: 13--17 years per seed. Upward arrows mark clipped bars.}
    \Description{Six bar charts, PEMS07 and TFNSW by MAE, RMSE, and MAPE, compare 21 configurations on newly added sensors. EvoST has the lowest MAE and RMSE on both networks, with 13.4 and 23.6 on PEMS07 and 85.2 and 182.5 on TFNSW. Frozen+RLS follows closely with 13.6 and 23.8 on PEMS07 and 85.4 and 183.0 on TFNSW. For reference, STAEFormer has 15.1 and 26.4 on PEMS07 and 137.0 and 272.2 on TFNSW.}
\else
\begin{figure}[t]
    \centering
    \includegraphics[width=\linewidth]{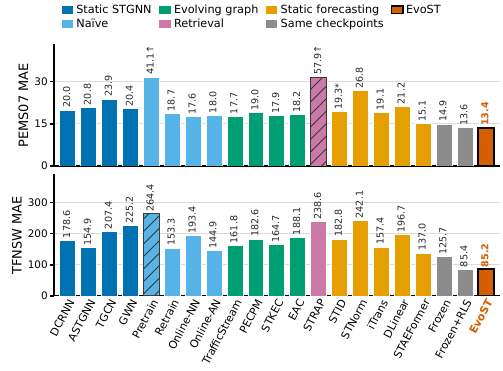}
    \caption{Sensors absent from the preceding year's graph (annual benchmark): H12 MAE averaged over 19 arrival years, then five seeds. RMSE and MAPE are in the online extended version. Frozen, Frozen+RLS, and EvoST use training-split input statistics. Other bars use split-specific statistics. ST-TTC is omitted: its archived new-sensor rows duplicate Retrain. *PEMS07 STID: 13--17 years per seed. Upward arrows mark clipped bars.}
    \Description{Two bar charts compare 21 configurations by MAE on newly added sensors of PEMS07 and TFNSW. EvoST has the lowest MAE on both networks (13.4 and 85.2), followed closely by Frozen+RLS (13.6 and 85.4). For reference, STAEFormer has 15.1 and 137.0.}
\fi
    \label{fig:new_sensors}
\ifextended
\end{figure*}
\else
\end{figure}
\fi

\ifextended\else
\AblationFigure
\SensitivityFigure
\fi

\paragraph{Overall performance.}
\Cref{tab:main_by_method_part1} reports three representative networks. EvoST
has the lowest MAE, RMSE, and MAPE among the listed non-foundation methods on all
three, and also on the other seven PEMS districts\extref{
(\Cref{tab:main_by_method_part2,tab:main_by_method_part3} in Appendix~\ref{app:ten_networks})}{, reported in the online extended version}.
\extref{The annual benchmark uses checkpoints trained for the same year. With these
checkpoints and training-split input normalization, EvoST reduces Avg-MAE
relative to the frozen forecaster by 7.2\%, 10.7\%, and 22.0\% on PEMS03,
PEMS04, and TFNSW (\Cref{fig:annual_controls}), and the matched
frozen-plus-RLS control already achieves most of this reduction. On all ten
networks, EvoST's Avg-MAE is below that of Frozen+RLS, by 0.8\% on average.
On the three networks shown, the reductions are 0.8\%, 0.3\%, and 0.6\%, with lower
cross-year means in five, three, and five of the five paired evaluation
seeds. On PEMS03, calibrating the refreshed model raises MAPE from 21.66 to
22.52.}{The annual benchmark uses checkpoints trained for the same year. Using them with training-split input normalization, EvoST reduces Avg-MAE relative to the frozen forecaster by 7.2\%, 10.7\%, and 22.0\% on PEMS03, PEMS04, and TFNSW (\Cref{fig:annual_controls}). The matched Frozen+RLS control achieves most of this reduction, and EvoST's Avg-MAE is below it on every network, by 0.8\% on average. \Cref{fig:ablation}(a) compares Refresh, Frozen+RLS, and EvoST against the same Frozen checkpoints.}

\paragraph{Newly added sensors.}
These annual-benchmark sensors were absent from the preceding year's graph
but were included in the current year's STAEFormer checkpoint. EvoST has the
lowest MAE and RMSE of the plotted configurations on PEMS07 and TFNSW
(\Cref{fig:new_sensors}\extref{}{, RMSE in the online extended version}).
Frozen+RLS MAE is at most 1.3\% higher. The difference between Frozen and
STAEFormer reflects input standardization.

\ifextended
\SensitivityFigure
\AblationFigure
\fi

\paragraph{Comparison with time-series foundation models.}
TSFMs are reference rows outside the non-foundation ranking. \extref{EvoST has lower
MAE and RMSE than every zero-shot and fine-tuned TSFM cell in
\Cref{tab:main_by_method_part1}, and lower MAPE except on PEMS04, where
Chronos-2 (FT) reaches 15.25 against 15.52.}{EvoST has lower MAE and RMSE than every TSFM cell in \Cref{tab:main_by_method_part1} and lower MAPE except Chronos-2 (FT) on PEMS04 (15.25 against 15.52).}\extref{ At horizons 3, 6, and 12 on all ten
networks (\Cref{tab:main_part1,tab:main_part2}), EvoST also has lower MAE and RMSE
than every TSFM reference. Appendix~\ref{app:ten_networks} lists the MAPE exceptions.}{}

\paragraph{Forecast horizons.}
\extref{On PEMS03 and TFNSW, at each of the 12 steps (\Cref{per_horizon}\extref{}{, Appendix~\ref{app:annual}}), the calibrator gives
most of the MAE and RMSE reduction relative to the frozen checkpoints, and
identity refresh lowers both further. On PEMS03, the MAE reduction of EvoST over
Frozen+RLS grows from 0.3\% at the first step to 1.5\%
at the last. Over all ten networks, EvoST is below Frozen in all 360
combinations of network, metric, and step and below Frozen+RLS in
343\extref{ (\Cref{per_horizon_all_app})}{}. Of the 17 exceptions, 14 are MAPE steps on PEMS04 and
PEMS05.}{Per step, the calibrator gives most of the MAE and RMSE reduction on PEMS03 and TFNSW (\Cref{per_horizon}, Appendix~\ref{app:annual}). Over the ten networks, EvoST is below Frozen in all 360 combinations of network, metric, and step and below Frozen+RLS in 343, with 14 of the 17 exceptions MAPE steps on PEMS04 and PEMS05.}

\newcommand{\PerStepFigure}[3]{%
\begin{figure}[#1]
    \centering
    \includegraphics[width=#2\linewidth]{#3}
    \caption{Per-step errors on all sensors of PEMS03 and TFNSW: EvoST and the same checkpoints with the calibrator only (Frozen+RLS) or without adaptation (Frozen), averaged over 25 and 21 years and five seeds.}
    \label{per_horizon}
    \Description{Six line plots compare MAE, RMSE, and MAPE of EvoST, Frozen+RLS, and Frozen over forecast steps 1 to 12 on PEMS03 and TFNSW.}
\end{figure}}
\ifextended
\PerStepFigure{h!}{0.99}{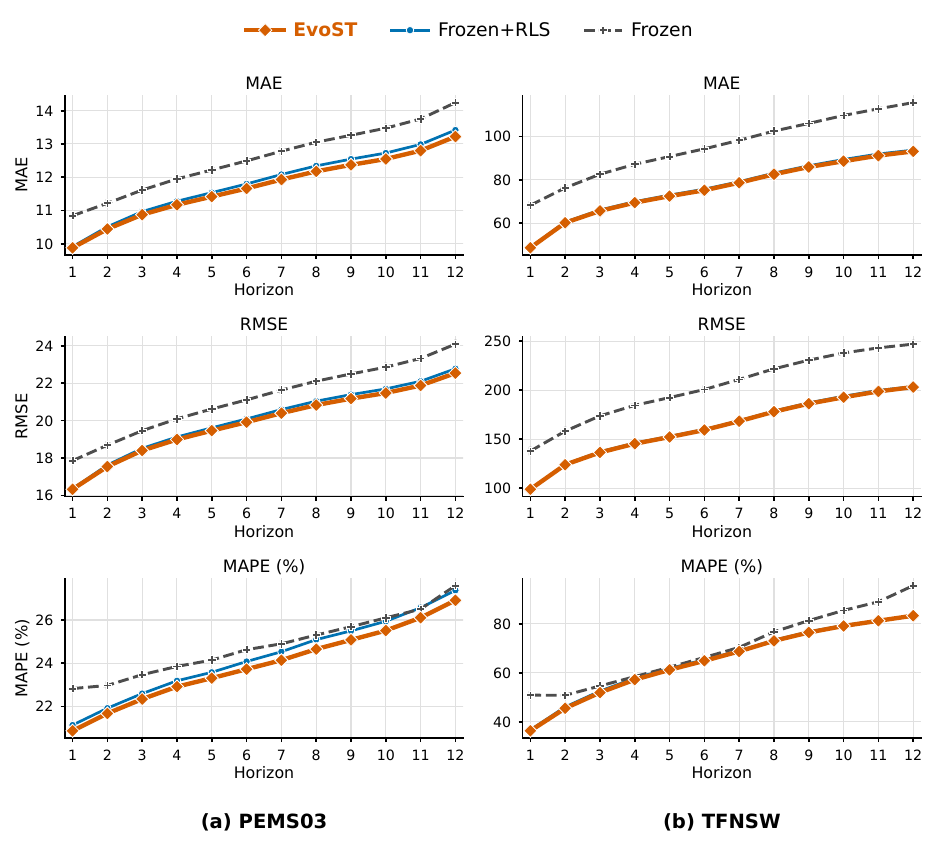}
\fi

\subsection{Ablation Study}
\label{sec:ablation}

\Cref{fig:ablation} separates the parts of EvoST. \extref{On the
annual benchmark (a), identity refresh alone lowers ten-network Avg-MAE
relative to the frozen checkpoints by 3.7\%, the calibrator alone by 8.7\%,
and both by 9.4\%. For Avg-MAPE, refresh alone gives the larger reduction
(7.8\% against 6.9\%). Without any calibrator (b), the old checkpoint's H12
MAE is 47.9\% above that of the uncalibrated target-year checkpoint, and the
refreshed checkpoint's is 1.5\% below it. With all three models calibrated,
the ratios are 1.177 and 1.025.}{On the annual benchmark (a), the calibrator gives most of the ten-network Avg-MAE reduction relative to the frozen checkpoints and identity refresh adds to it, while for Avg-MAPE refresh alone gives the larger reduction. Without any calibrator (b), the old checkpoint's H12 MAE is 47.9\% above that of the uncalibrated target-year checkpoint and the refreshed checkpoint's is 1.5\% below it.}

\subsection{Sensitivity Analysis}
\label{sec:sensitivity}

\Cref{fig:sensitivity} varies EvoST's settings. \extref{On the
tested annual validation-selection grid (a), the ten-network mean error
varies more across forgetting factors than across ridge penalties: at
$\lambda=0.99$, the mean excess H12 MAE above each run's best setting is
5.7\% for $\rho=1$ and 4.1\% for $\rho=100$, and at $\lambda\in\{0.999,1\}$
both means stay below 1\%. The lowest identity learning rate, $10^{-3}$, was
selected in 1,100 of the 1,160 annual fits (b), and $10^{-2}$ has a
ten-network mean excess holdout MAE of 4.0\% above each fit's best
candidate. Refresh needs enough recent history (c\extref{, \Cref{tab:budget}}{}, Appendix~\ref{app:evost_implementation}).
The budget sets both the fitting windows and the calibrator's selection
prefix. With 1,024 fitting windows, about 3.6 days of 5-minute data, refresh
lowers the calibrated ratio from 1.175 to 1.046, and with 2,048 to 1.026.
Without calibration, it lowers MAE at every tested budget, by 8\% with 64
windows and by 33\% with 2,048. \Cref{sec:limitations} covers the calibrated
small budgets. Refresh also stays effective for older
checkpoints (d): at ages of five, ten, and fifteen years, it recovers 91\%,
85\%, and 89\% of gaps of 18.3\%, 21.7\%, and 21.2\% (ratios 1.016, 1.033,
and 1.023).}{On the annual validation-selection grid (a), the ten-network mean error varies more across forgetting factors than across ridge penalties, and the mean excess over each run's best setting stays below 1\% for $\lambda\in\{0.999,1\}$. The lowest identity learning rate, $10^{-3}$, was selected in 1,100 of the 1,160 annual fits (b). Refresh needs enough recent history (c, Appendix~\ref{app:evost_implementation}). The budget varies fitting windows and the calibrator's selection prefix together. With 1,024 fitting windows, about 3.6 days of 5-minute data, refresh lowers the calibrated ratio from 1.175 to 1.046, and without calibration it lowers MAE at every tested budget.}

\ifextended
\begin{table}[t]
\centering
\caption{History budgets on twelve region-year settings (six districts, 2020 and 2025). Calibrated test H12 MAE ratios to the calibrated target-year checkpoint at the same budget are geometric means after averaging three seeds per setting. Fit windows include the identity holdout. Selection windows are an additional validation prefix for calibration. The identity learning rate is fixed at $3\times10^{-3}$.}
\label{tab:budget}
\begin{tabular}{rrrr}
\toprule
Fit windows & Selection & Old & Refresh \\
\midrule
64 & 32 & 1.208 & 1.214 \\
256 & 128 & 1.153 & 1.198 \\
1,024 & 512 & 1.175 & 1.046 \\
2,048 & 1,024 & 1.177 & 1.026 \\
\bottomrule
\end{tabular}
\end{table}

\fi

\subsection{Case Study}
\label{sec:case}

\begin{figure}[t]
    \centering
    \includegraphics[width=\linewidth]{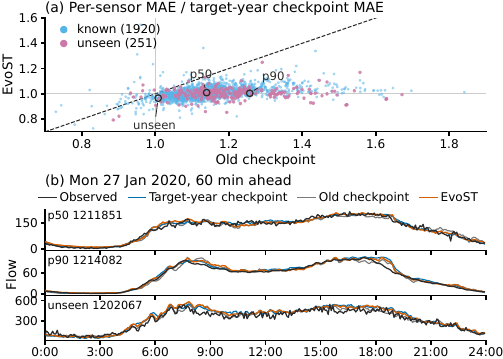}
    \caption{Calibrated case study on PEMS12/2020. Panel (a) shows per-sensor H12 MAE ratios to the target-year checkpoint, and (b) 60-minute-ahead forecasts. See Appendix~\ref{app:evost_implementation}.}
    \label{fig:case}
    \Description{Panel (a) plots 2,171 sensors with the old checkpoint's MAE ratio on the horizontal axis and EvoST's on the vertical axis. Most points lie near 1.0 vertically and between 1.0 and 1.4 horizontally. Panel (b) shows observed flow and the 60-minute-ahead forecasts of the target-year checkpoint, the old checkpoint, and EvoST for three sensors over one day.}
\end{figure}

\Cref{fig:case} examines PEMS12/2020. For sensors with valid test targets, the
median ratio of calibrated test H12 MAE to the target-year checkpoint falls
from 1.13 to 1.01 for the 1,920 the old checkpoint has seen and from 1.18 to
1.00 for the 251 it has not, and EvoST beats the old checkpoint on 97\% of
them (a). On 27 January (b), EvoST's mean absolute differences from the
target-year forecasts are 5.4, 2.5, and 14.1, versus 8.3, 3.9, and 25.8 for
the old checkpoint.

\section{Limitations}
\label{sec:limitations}

\extref{EvoST relies on labels. Identity refresh uses recent windows whose targets
have been released, and the calibrator learns from delayed feedback. The
method therefore does not cover label-free deployment. With short histories, EvoST can
be worse than calibrating the old checkpoint alone: at the two smallest
budgets in \Cref{fig:sensitivity}(c), the calibrated old checkpoint has lower MAE
than the calibrated refreshed model.}{EvoST relies on labels: identity refresh uses recent windows whose targets have been released and the calibrator learns from delayed feedback, so label-free deployment is not covered. With short histories, calibrating the old checkpoint alone can be better (the two smallest budgets in \Cref{fig:sensitivity}(c)).} \extref{Identities are carried across snapshots
by sensor ID. The arrival study provides 64 post-arrival history windows
before identity borrowing and does not evaluate commissioning without such
history. EvoST applies to backbones whose node-specific parameters can
be separated from the shared ones, and we evaluate three such backbones.
Identity fitting still backpropagates through the frozen backbone, so its
time and memory depend on the backbone and the graph size. The evaluation
covers univariate traffic flow with yearly snapshots. In the annual
benchmark, where each year has its own checkpoint, most of the Avg-MAE
reduction relative to the frozen checkpoint comes from the calibrator.}{Identities are carried across snapshots by sensor ID, and arriving sensors receive 64 history windows before identity borrowing. Commissioning without such history is not evaluated. EvoST needs backbones whose node-specific parameters can be separated from the shared ones (we evaluate three), and identity fitting still backpropagates through the frozen backbone. The evaluation covers univariate traffic flow with yearly snapshots, and in the annual benchmark most of the Avg-MAE reduction relative to the frozen checkpoint comes from the calibrator.}

\section{Conclusion}

We studied how to maintain a deployed traffic forecaster while its sensor
graph evolves. \extref{EvoST keeps the shared spatio-temporal weights of a checkpoint
fixed and refreshes node identities from released labels. New sensors borrow
identities from behaviorally similar established sensors until a joint
refit, and per-sensor recursive least squares corrects forecasts as feedback
arrives.}{EvoST keeps a checkpoint's shared spatio-temporal weights fixed, refreshes node identities from released labels, lets new sensors borrow identities from behaviorally similar established sensors until a joint refit, and corrects forecasts with per-sensor recursive least squares as feedback arrives.} \extref{With all compared forecasters calibrated, identity refresh recovers
86\% of the test H12 MAE gap between five-year-old and target-year STAEFormer
checkpoints on twelve PeMS region-year settings. Its MAE is 0.7\% above full
fine-tuning on the same recent data. Similar recovery holds for checkpoints
up to fifteen years old, and on sixteen settings the refresh recovers 67\%
and 88\% of the five-year gaps for STID and Graph WaveNet. On the annual
benchmark of ten networks, EvoST achieves the lowest Avg-MAE, Avg-RMSE, and
Avg-MAPE among the reported non-foundation methods.}{With all compared forecasters calibrated, identity refresh recovers 86\% of the test H12 MAE gap between five-year-old and target-year STAEFormer checkpoints on twelve PeMS region-year settings, with MAE 0.7\% above full fine-tuning on the same recent data. Similar recovery holds for checkpoints up to fifteen years old, and the refresh recovers 67\% and 88\% of the five-year gaps for STID and Graph WaveNet. On the annual benchmark of ten networks, EvoST achieves the lowest Avg-MAE, Avg-RMSE, and Avg-MAPE among the reported non-foundation methods.} In the evaluated architectures and protocol, node identities are an
effective maintenance target: refreshing them recovers most of the measured
degradation, and shared weights remain largely reusable with calibration.

\begin{acks}
This work was supported by the ARC Centre of Excellence for Automated Decision-Making and Society (CE200100005). We acknowledge the resources and services provided by the National Computational Infrastructure (NCI), which is supported by the Australian Government. This research is also partially supported by the ARC Training Centre for Whole Life Design of Carbon Neutral Infrastructure (IC230100015).
\end{acks}

\ifextended\else\newpage\fi
\bibliographystyle{ACM-Reference-Format}
\bibliography{acm}

\appendix

\setcounter{topnumber}{3}
\setcounter{bottomnumber}{2}
\setcounter{totalnumber}{5}
\setcounter{dbltopnumber}{2}
\renewcommand{\topfraction}{0.95}
\renewcommand{\bottomfraction}{0.90}
\renewcommand{\textfraction}{0.05}
\renewcommand{\floatpagefraction}{0.82}
\renewcommand{\dbltopfraction}{0.95}
\renewcommand{\dblfloatpagefraction}{0.82}
\setlength{\dbltextfloatsep}{10pt plus 2pt minus 2pt}

\section{Appendix}
\label{appendix}

\subsection{Dataset Scale and Structural Evolution}
\label{app:structural_evolution}

\begin{figure}[t]
    \centering
    \includegraphics[width=0.99\linewidth]{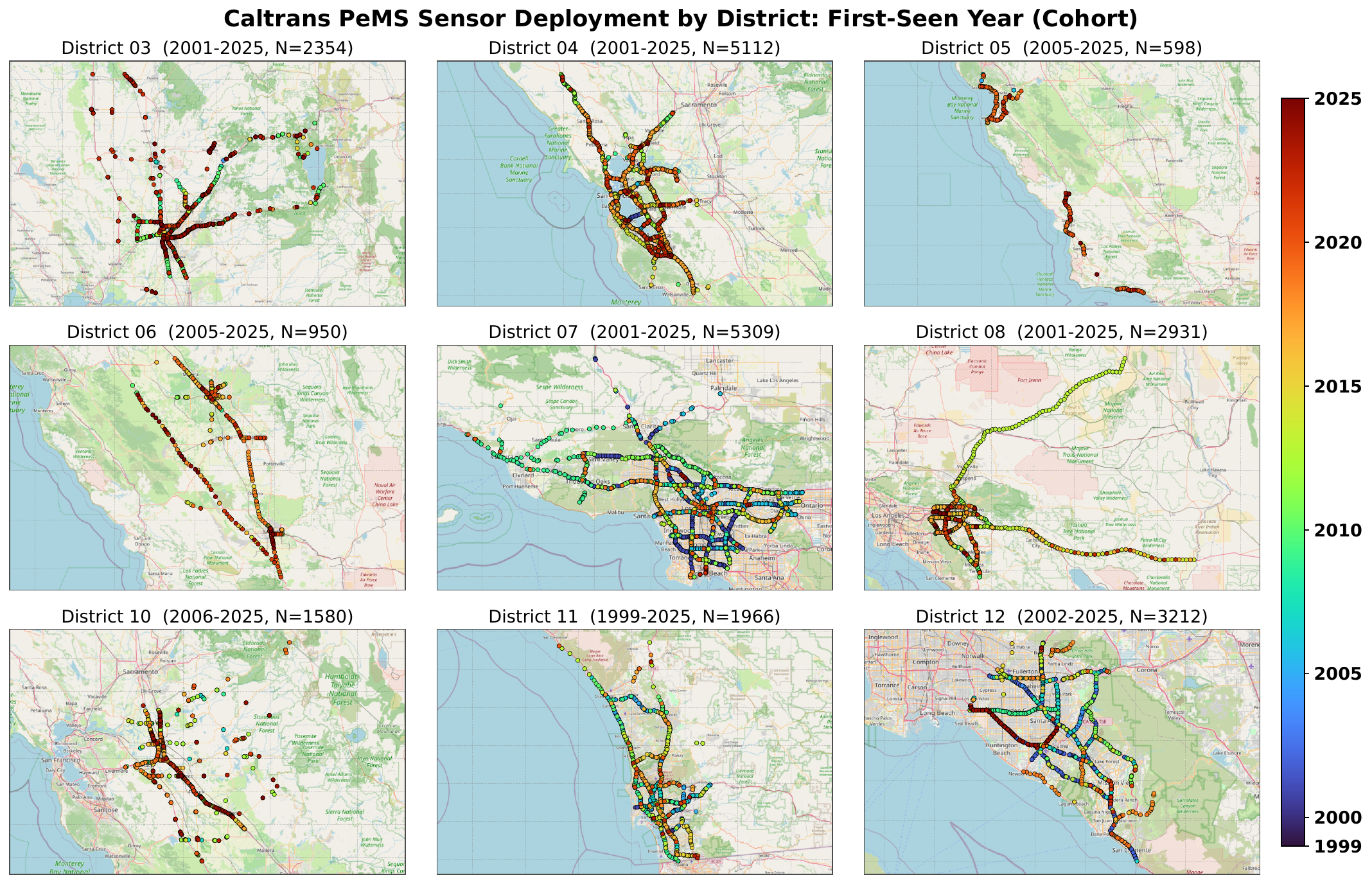}
    \caption{Sensor coverage across nine districts, colored by first appearance in
    the yearly snapshots. $N$ counts all sensors with valid coordinates in the
    displayed years.}
    \label{fig:evo_deployment}
    \Description{Nine maps color sensors by their first appearance in the
    yearly dataset snapshots. Each panel reports the cumulative number of
    mapped sensors across its displayed years.}
\end{figure}

\begin{figure}[!h]
\centering
\includegraphics[width=0.99\linewidth]{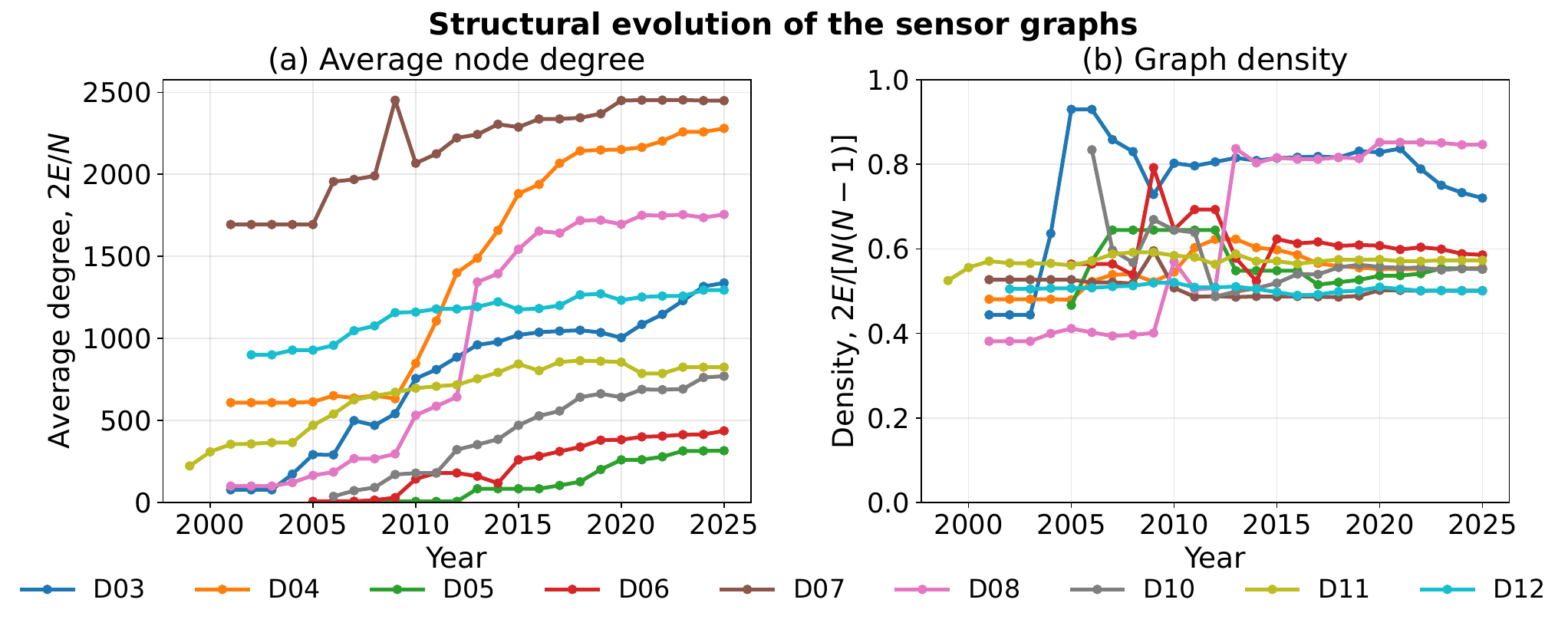}
\caption{Yearly average node degree and density of the sensor proximity
graphs. Both vary non-monotonically.}
\label{fig:graph_evo}
\Description{Two groups of yearly line charts show non-monotonic changes in
average node degree and graph density across the nine EvoXXLTraffic
districts.}
\end{figure}

The nine PeMS districts cover 1999--2025 with 22.1 million five-minute
timesteps and 36.0 billion observations (144\,GB, \Cref{tab:evo_scale}).
\Cref{fig:evo_deployment} colors every mapped sensor by the first yearly
snapshot in which it appears: later cohorts both extend coverage along new
corridors and densify existing ones. \Cref{fig:graph_evo} tracks the average
node degree $2E/N$ and the density $2E/[N(N-1)]$ of the yearly proximity
graphs, with $N$ nodes and $E$ undirected edges counted once. Both change
non-monotonically as sensors are added and removed, so successive snapshots
differ in size and connectivity.

\suppressfloats[t]
\begin{table}[t]
\centering
\footnotesize
\caption{\textsc{EvoXXLTraffic} scale and sensor growth.}
\label{tab:evo_scale}
\resizebox{\linewidth}{!}{%
\begin{tabular}{lccccrr}
\toprule
District & Span & \#Yr & $N_{\text{first}}\!\to\!N_{\text{last}}$ & Timesteps & Obs. & Size \\
\midrule
D03 & 2001--2025 & 25 & 174$\to$1859 & 2,611,872 & \textcolor{obsblue}{2.56B} & \textcolor{sizewine}{10.2\,GB} \\
D04 & 2001--2025 & 25 & 1266$\to$4110 & 2,559,744 & \textcolor{obsblue}{6.63B} & \textcolor{sizewine}{26.5\,GB} \\
D05 & 2005--2025 & 21 & 6$\to$572 & 2,185,344 & \textcolor{obsblue}{0.50B} & \textcolor{sizewine}{2.0\,GB} \\
D06 & 2005--2025 & 21 & 13$\to$746 & 2,138,976 & \textcolor{obsblue}{0.84B} & \textcolor{sizewine}{3.4\,GB} \\
D07 & 2001--2025 & 25 & 3215$\to$4888 & 2,629,728 & \textcolor{obsblue}{11.28B} & \textcolor{sizewine}{45.1\,GB} \\
D08 & 2001--2025 & 25 & 262$\to$2074 & 2,612,736 & \textcolor{obsblue}{3.49B} & \textcolor{sizewine}{14.0\,GB} \\
D10 & 2006--2025 & 20 & 45$\to$1396 & 2,051,712 & \textcolor{obsblue}{1.70B} & \textcolor{sizewine}{6.8\,GB} \\
D11 & 1999--2025 & 27 & 424$\to$1440 & 2,788,128 & \textcolor{obsblue}{3.23B} & \textcolor{sizewine}{12.9\,GB} \\
D12 & 2002--2025 & 24 & 1783$\to$2587 & 2,524,608 & \textcolor{obsblue}{5.73B} & \textcolor{sizewine}{22.9\,GB} \\
\midrule
\textbf{Total} & 1999--2025 & -- & -- & 22,102,848 & \textbf{\textcolor{obsblue}{35.97B}} & \textbf{\textcolor{sizewine}{144\,GB}} \\
\bottomrule
\end{tabular}%
}
\end{table}

\ifextended
\subsection{Extended Related Work}
\label{app:extended_related_work}

\subsubsection{Spatio-Temporal Traffic Forecasting}

Traffic forecasting, including traffic flow, speed, and congestion status, has always been one of the most important issues in computational urban and intelligent transportation systems. Early traffic forecasting methods primarily relied on statistical modeling and traditional machine learning methods, such as Historical Average (HA), Autoregressive Integrated Moving Average (ARIMA)~\cite{box2015time}, Seasonal ARIMA~\cite{williams2003modeling}, Kalman filtering~\cite{kumar2017traffic}, Vector Autoregression (VAR)~\cite{schimbinschi2017topology}, Support Vector Regression (SVR)~\cite{wu2004travel}, and $k$-Nearest Neighbors (kNN)~\cite{cai2016spatiotemporal}. These methods typically treat traffic sequences as one-dimensional or low-dimensional time series and make predictions by modeling trends, periodicity, and short-term autocorrelation in historical observations. Among them, HA uses historical averages over the same period as predictions, making it simple and efficient but difficult to adapt to sudden changes. ARIMA and its variants can characterize linear temporal dependence and periodic patterns but usually rely on stationarity assumptions. Kalman filtering is suitable for online state estimation and short-term forecasting, while SVR and kNN improve predictive capability through nonlinear mapping or similar-pattern matching. Nevertheless, these methods mostly rely on manual feature design and struggle to characterize complex nonlinear temporal variations and spatial dependencies within road networks simultaneously.

Deep traffic forecasting has used convolutional neural networks (CNNs), recurrent neural networks (RNNs), graph neural networks (GNNs), Transformers, and, more recently, state-space models. Early temporal models used long short-term memory (LSTM) or gated recurrent unit (GRU) networks to capture long-range dependencies. Temporal models also combine recurrence with convolution-based pattern attention~\cite{shih2019temporal}. Related sequence-modeling components include the dilated causal convolutions used by WaveNet for autoregressive audio modeling~\cite{van2016wavenet}. Graph models then made road-network dependencies explicit. Representative methods include T-GCN~\cite{zhao2019t}, STGCN~\cite{yu2018spatio}, and DCRNN~\cite{li2018diffusion}. T-GCN places graph convolution inside GRU units, STGCN combines graph and temporal convolutions, and DCRNN models directed propagation with diffusion convolution and an encoder-decoder architecture. GNNs have also been applied to regional multimodal demand forecasting: Ma et al.~\cite{ma2022forecasting} release an NYC taxi-bike dataset and study multi-source, multi-graph, and meta-information-enhanced STGNNs. Our setting instead concerns sensor graphs whose nodes and connectivity change over time.

Traffic forecasting models have since developed toward adaptive graph learning, attention mechanisms, and large-scale sequence modeling. Graph WaveNet~\cite{wu2019graph}, for example, reduces reliance on predefined adjacency matrices by learning adaptive spatial dependencies. ASTGCN~\cite{guo2019attention}, GMAN~\cite{zheng2020gman}, and STTN~\cite{xu2020spatial} introduce spatio-temporal attention to dynamically model relationships among sensors and time steps. Transformer-based models are now widely used because of their long-sequence modeling capability, while state-space models offer linear-complexity alternatives for large-scale forecasting. This development has produced many strong models, including MTMGNN~\cite{yin2023mtmgnn}, STQCL~\cite{yin2024enhancing}, STAEFormer~\cite{liu2023spatio}, STID~\cite{shao2022spatial}, HIMNet~\cite{dong2024heterogeneity}, PDFormer~\cite{jiang2023pdformer}, and AutoSTF~\cite{lyu2025autostf}.

Despite their progress on static benchmarks, most existing methods assume that the sensor set and road-network topology remain unchanged between training and testing. In real deployments, however, sensor networks expand over time, new sensors are commissioned, and connectivity and traffic patterns drift. A deployed forecaster therefore faces not merely a longer time series but a continuously evolving traffic graph. The resulting node growth, topology change, and temporal distribution shift challenge models built under a fixed-graph assumption.

\subsubsection{Online and Continual Traffic Forecasting}

Long-term traffic-system deployment has consequently motivated online and continual forecasting. New sensors are repeatedly added, while traffic patterns at existing nodes evolve over time. An effective model must therefore acquire new spatio-temporal patterns without catastrophically forgetting previously learned knowledge.

TrafficStream~\cite{ijcai2021p0498} was among the first methods to formalize network expansion and pattern evolution as a streaming continual-learning problem. It uses local subgraphs to integrate neighborhood information for new sensors, Jensen--Shannon divergence to detect drift at existing nodes, and historical replay with parameter smoothing to mitigate catastrophic forgetting. Following an explicit-memory approach, PECPM~\cite{wang2023pattern} maintains a library of representative spatio-temporal patterns and expands and consolidates it as the graph evolves. STKEC~\cite{wang2023knowledge} selects influential old nodes to construct local subgraphs, combines them with a memory bank for long-term periodic patterns, and preserves prior knowledge through parameter-distance regularization. To reduce update cost, EAC~\cite{chen2025expand} uses a continual prompt pool to represent knowledge from different stages and follows an expansion--compression strategy that avoids large-scale updates of the whole model. STBP~\cite{liugeneral} freezes a general spatio-temporal backbone and expands only a scalable contextual pattern library, balancing stability and plasticity. STEV~\cite{ma2025beyond} treats new sensors as a variable-expansion problem and uses a flattening scheme to improve robustness to varying input scales. STRAP~\cite{zhang2025strap} constructs an external pattern library and retrieves similar patterns as prompts at inference time. ST-TTC~\cite{chen2025learning} performs test-time calibration with frequency-domain alignment and streaming memory queues, but primarily addresses temporal drift within a fixed node set.

Unlike methods that update substantial model components or depend on large replay memories, EvoST keeps the shared spatio-temporal weights of a deployed checkpoint fixed. It refreshes only the node identities from released labels, lets new sensors borrow identities from behaviorally similar established sensors until a joint refit, and corrects the forecasts of each sensor by recursive least squares as delayed feedback arrives.

\subsection{Annual Benchmark Results on All Ten Networks}
\label{app:ten_networks}

\providecommand{\best}[1]{\cellcolor{yellow!35}\textbf{\textcolor{red}{#1}}}
\providecommand{\second}[1]{\cellcolor{black!10}\underline{#1}}
\begin{table*}[!h]
\centering
\caption{Average results on PEMS05--08 (mean$\pm$standard deviation, with the sample standard deviation over five seeds for EvoST).}
\label{tab:main_by_method_part2}
\setlength{\tabcolsep}{2.2pt}
\scriptsize
\renewcommand{\arraystretch}{0.85}
\resizebox{\textwidth}{!}{%
\begin{tabular}{ll|ccc|ccc|ccc|ccc}
\toprule
\multirow{2}{*}{\textbf{Category}} & \multirow{2}{*}{\textbf{Baseline}} & \multicolumn{3}{c|}{\textbf{PEMS05}} & \multicolumn{3}{c|}{\textbf{PEMS06}} & \multicolumn{3}{c|}{\textbf{PEMS07}} & \multicolumn{3}{c}{\textbf{PEMS08}} \\
\cmidrule(lr){3-5} \cmidrule(lr){6-8} \cmidrule(lr){9-11} \cmidrule(lr){12-14}
 & & \textbf{MAE} & \textbf{RMSE} & \textbf{MAPE (\%)} & \textbf{MAE} & \textbf{RMSE} & \textbf{MAPE (\%)} & \textbf{MAE} & \textbf{RMSE} & \textbf{MAPE (\%)} & \textbf{MAE} & \textbf{RMSE} & \textbf{MAPE (\%)} \\
\midrule
\multirow{4}{*}{\textbf{\shortstack[l]{Static\\STGNN\\Backbones}}} & DCRNN & 11.69$_{\pm0.07}$ & 18.34$_{\pm0.09}$ & 26.98$_{\pm0.74}$ & 14.87$_{\pm0.31}$ & 23.29$_{\pm0.43}$ & 24.66$_{\pm0.77}$ & 19.39$_{\pm0.28}$ & 32.62$_{\pm0.30}$ & 31.09$_{\pm1.27}$ & 15.50$_{\pm0.13}$ & 25.20$_{\pm0.19}$ & 23.31$_{\pm0.41}$ \\
 & ASTGNN & 11.69$_{\pm0.04}$ & 18.52$_{\pm0.08}$ & 25.31$_{\pm0.60}$ & 15.52$_{\pm0.16}$ & 24.04$_{\pm0.28}$ & 23.41$_{\pm0.46}$ & 19.56$_{\pm0.05}$ & 33.05$_{\pm0.13}$ & 29.95$_{\pm0.26}$ & 15.22$_{\pm0.05}$ & 25.23$_{\pm0.10}$ & 20.85$_{\pm0.39}$ \\
 & TGCN & 11.94$_{\pm0.10}$ & 19.43$_{\pm0.16}$ & 26.83$_{\pm0.59}$ & 15.52$_{\pm0.46}$ & 25.38$_{\pm0.56}$ & 24.06$_{\pm0.40}$ & 20.62$_{\pm0.41}$ & 37.75$_{\pm0.75}$ & 32.43$_{\pm1.02}$ & 15.86$_{\pm0.17}$ & 27.85$_{\pm0.40}$ & 22.13$_{\pm0.71}$ \\
 & GWN & 12.57$_{\pm0.13}$ & 19.49$_{\pm0.18}$ & 27.19$_{\pm0.86}$ & 16.16$_{\pm0.54}$ & 24.84$_{\pm0.65}$ & 24.76$_{\pm0.81}$ & 19.51$_{\pm0.99}$ & 32.22$_{\pm1.35}$ & 31.81$_{\pm2.36}$ & 15.69$_{\pm0.16}$ & 25.27$_{\pm0.29}$ & 21.90$_{\pm0.74}$ \\
\midrule
\multirow{4}{*}{\textbf{\shortstack[l]{Na\"ive \\Schemes}}} & Pretrain & 24.94$_{\pm0.99}$ & 34.77$_{\pm1.06}$ & 116.10$_{\pm7.29}$ & 78.66$_{\pm1.39}$ & 97.96$_{\pm2.03}$ & 79.40$_{\pm1.33}$ & 40.30$_{\pm2.05}$ & 54.10$_{\pm1.93}$ & 49.24$_{\pm2.54}$ & 52.25$_{\pm2.71}$ & 62.21$_{\pm3.13}$ & 57.09$_{\pm1.58}$ \\
 & Retrain & 11.31$_{\pm0.04}$ & 18.03$_{\pm0.07}$ & 24.42$_{\pm0.41}$ & 14.37$_{\pm0.34}$ & 22.39$_{\pm0.39}$ & 22.54$_{\pm0.33}$ & 18.02$_{\pm0.27}$ & 30.73$_{\pm0.23}$ & 27.64$_{\pm1.35}$ & 14.34$_{\pm0.07}$ & 23.85$_{\pm0.09}$ & 20.58$_{\pm0.25}$ \\
 & Online-NN & 15.98$_{\pm0.52}$ & 26.35$_{\pm0.67}$ & 29.06$_{\pm0.76}$ & 27.64$_{\pm0.65}$ & 38.29$_{\pm1.20}$ & 32.16$_{\pm0.38}$ & 22.72$_{\pm0.59}$ & 35.82$_{\pm0.68}$ & 35.08$_{\pm1.56}$ & 17.30$_{\pm0.05}$ & 26.86$_{\pm0.13}$ & 24.09$_{\pm0.35}$ \\
 & Online-AN & 11.10$_{\pm0.07}$ & 17.63$_{\pm0.13}$ & 23.49$_{\pm0.44}$ & 14.02$_{\pm0.21}$ & 21.68$_{\pm0.28}$ & 21.93$_{\pm0.52}$ & 17.67$_{\pm0.31}$ & 30.13$_{\pm0.32}$ & 26.68$_{\pm1.27}$ & 13.92$_{\pm0.10}$ & 23.20$_{\pm0.15}$ & 19.28$_{\pm0.22}$ \\
\midrule
\multirow{4}{*}{\textbf{\shortstack[l]{Evolving\\Graph}}} & TrafficStream & 11.32$_{\pm0.13}$ & 18.04$_{\pm0.14}$ & 24.56$_{\pm0.93}$ & 14.46$_{\pm0.34}$ & 22.41$_{\pm0.47}$ & 22.43$_{\pm0.49}$ & 17.54$_{\pm0.31}$ & 30.04$_{\pm0.29}$ & 26.67$_{\pm2.19}$ & 14.04$_{\pm0.13}$ & 23.35$_{\pm0.18}$ & 20.96$_{\pm1.33}$ \\
 & PECPM & 11.67$_{\pm0.31}$ & 18.50$_{\pm0.44}$ & 25.13$_{\pm0.85}$ & 18.02$_{\pm3.19}$ & 29.03$_{\pm6.07}$ & 26.88$_{\pm5.50}$ & 19.23$_{\pm1.47}$ & 32.52$_{\pm2.46}$ & 27.85$_{\pm1.72}$ & 15.58$_{\pm0.77}$ & 25.25$_{\pm1.04}$ & 22.26$_{\pm1.64}$ \\
 & STKEC & 11.48$_{\pm0.16}$ & 18.26$_{\pm0.21}$ & 25.11$_{\pm0.78}$ & 14.90$_{\pm0.24}$ & 23.19$_{\pm0.34}$ & 24.14$_{\pm0.58}$ & 17.07$_{\pm1.26}$ & 29.33$_{\pm1.84}$ & 25.66$_{\pm3.04}$ & 14.47$_{\pm0.38}$ & 23.89$_{\pm0.50}$ & 21.55$_{\pm1.34}$ \\
 & EAC & 16.35$_{\pm3.53}$ & 23.77$_{\pm3.62}$ & 61.09$_{\pm37.33}$ & 21.10$_{\pm4.19}$ & 31.23$_{\pm4.99}$ & 36.63$_{\pm16.01}$ & 18.26$_{\pm0.27}$ & 30.41$_{\pm0.37}$ & 28.73$_{\pm0.39}$ & 14.76$_{\pm0.28}$ & 23.76$_{\pm0.37}$ & 26.61$_{\pm1.73}$ \\
\midrule
\multirow{2}{*}{\textbf{\shortstack[l]{Retrieval\\TTC}}} & STRAP & 11.33$_{\pm0.04}$ & 17.97$_{\pm0.04}$ & 25.25$_{\pm0.88}$ & 14.44$_{\pm0.46}$ & 22.53$_{\pm0.49}$ & 22.60$_{\pm0.66}$ & 17.81$_{\pm0.16}$ & 30.37$_{\pm0.18}$ & 26.88$_{\pm0.66}$ & 14.15$_{\pm0.03}$ & 23.54$_{\pm0.04}$ & 19.96$_{\pm0.41}$ \\
 & ST-TTC & 11.22$_{\pm0.04}$ & 17.89$_{\pm0.07}$ & 24.33$_{\pm0.41}$ & 14.25$_{\pm0.34}$ & 22.21$_{\pm0.38}$ & 22.42$_{\pm0.33}$ & 17.74$_{\pm0.27}$ & 30.37$_{\pm0.23}$ & 27.40$_{\pm1.33}$ & 14.16$_{\pm0.07}$ & 23.63$_{\pm0.08}$ & 20.45$_{\pm0.25}$ \\
\midrule
\multirow{5}{*}{\textbf{\shortstack[l]{Static\\Forecasting\\Backbones}}} & STID & 11.64$_{\pm0.06}$ & 18.69$_{\pm0.08}$ & 24.51$_{\pm0.52}$ & 15.39$_{\pm0.23}$ & 24.34$_{\pm0.40}$ & 22.72$_{\pm0.38}$ & 19.27$_{\pm0.19}$ & 32.22$_{\pm0.21}$ & 34.13$_{\pm0.84}$ & 15.12$_{\pm0.07}$ & 25.69$_{\pm0.17}$ & 20.33$_{\pm0.76}$ \\
 & STNorm & 12.57$_{\pm0.10}$ & 19.00$_{\pm0.17}$ & 35.42$_{\pm0.64}$ & 16.70$_{\pm0.35}$ & 24.97$_{\pm0.45}$ & 32.58$_{\pm1.07}$ & 25.17$_{\pm1.15}$ & 39.57$_{\pm1.56}$ & 58.70$_{\pm3.03}$ & 18.33$_{\pm0.31}$ & 28.00$_{\pm0.33}$ & 43.33$_{\pm2.01}$ \\
 & iTrans & 11.42$_{\pm0.06}$ & 18.04$_{\pm0.07}$ & 24.92$_{\pm0.37}$ & 15.18$_{\pm0.41}$ & 23.39$_{\pm0.46}$ & 24.00$_{\pm0.96}$ & 18.64$_{\pm0.09}$ & 31.19$_{\pm0.17}$ & 30.76$_{\pm1.13}$ & 14.60$_{\pm0.08}$ & 23.99$_{\pm0.13}$ & 22.07$_{\pm0.99}$ \\
 & DLinear & 12.16$_{\pm0.05}$ & 19.40$_{\pm0.06}$ & 26.15$_{\pm0.70}$ & 16.40$_{\pm0.13}$ & 25.55$_{\pm0.15}$ & 24.49$_{\pm0.37}$ & 19.85$_{\pm0.01}$ & 33.90$_{\pm0.07}$ & 29.94$_{\pm0.50}$ & 15.58$_{\pm0.04}$ & 26.07$_{\pm0.04}$ & 21.16$_{\pm0.42}$ \\
 & STAEFormer & \second{10.16$_{\pm0.07}$} & \second{16.07$_{\pm0.10}$} & \second{22.83$_{\pm0.44}$} & \second{13.13$_{\pm0.60}$} & \second{19.88$_{\pm0.68}$} & \second{20.28$_{\pm0.85}$} & \second{15.32$_{\pm0.28}$} & \second{26.93$_{\pm0.26}$} & \second{21.39$_{\pm0.65}$} & \second{12.52$_{\pm0.03}$} & \second{20.88$_{\pm0.09}$} & \second{17.66$_{\pm0.23}$} \\
\midrule
\multirow{6}{*}{\textbf{\shortstack[l]{Times-series\\Forecasting\\Foundation\\Model}}} & Chronos-2 (ZS) & 12.67$_{\pm0.00}$ & 20.54$_{\pm0.00}$ & 25.64$_{\pm0.00}$ & 15.59$_{\pm0.00}$ & 25.65$_{\pm0.00}$ & 22.27$_{\pm0.00}$ & 18.67$_{\pm0.00}$ & 34.18$_{\pm0.00}$ & 22.66$_{\pm0.00}$ & 16.19$_{\pm0.00}$ & 27.58$_{\pm0.00}$ & 20.06$_{\pm0.00}$ \\
 & Chronos-2 (FT) & 11.12$_{\pm0.01}$ & 18.61$_{\pm0.02}$ & 21.67$_{\pm0.02}$ & 13.66$_{\pm0.01}$ & 23.12$_{\pm0.04}$ & 19.42$_{\pm0.04}$ & 16.63$_{\pm0.04}$ & 31.15$_{\pm0.13}$ & 20.04$_{\pm0.03}$ & 14.22$_{\pm0.02}$ & 25.08$_{\pm0.08}$ & 17.49$_{\pm0.03}$ \\
 & TimesFM2.5 (ZS) & 13.11$_{\pm0.00}$ & 21.07$_{\pm0.00}$ & 26.75$_{\pm0.00}$ & 16.08$_{\pm0.00}$ & 26.22$_{\pm0.00}$ & 22.96$_{\pm0.00}$ & 19.55$_{\pm0.00}$ & 35.34$_{\pm0.00}$ & 23.68$_{\pm0.00}$ & 16.87$_{\pm0.00}$ & 28.38$_{\pm0.00}$ & 20.96$_{\pm0.00}$ \\
 & TimesFM2.5 (FT) & 11.85$_{\pm0.01}$ & 19.92$_{\pm0.01}$ & 22.50$_{\pm0.02}$ & 14.28$_{\pm0.01}$ & 24.21$_{\pm0.01}$ & 20.06$_{\pm0.01}$ & 17.86$_{\pm0.02}$ & 33.57$_{\pm0.03}$ & 21.24$_{\pm0.02}$ & 14.90$_{\pm0.01}$ & 26.18$_{\pm0.02}$ & 18.26$_{\pm0.00}$ \\
 & Moirai-2.0 (ZS) & 13.26$_{\pm0.00}$ & 21.14$_{\pm0.00}$ & 25.93$_{\pm0.00}$ & 16.33$_{\pm0.00}$ & 26.23$_{\pm0.00}$ & 22.46$_{\pm0.00}$ & 20.58$_{\pm0.00}$ & 36.42$_{\pm0.00}$ & 23.57$_{\pm0.00}$ & 17.64$_{\pm0.00}$ & 29.31$_{\pm0.00}$ & 20.70$_{\pm0.00}$ \\
 & Moirai-2.0 (FT) & 11.66$_{\pm0.00}$ & 19.31$_{\pm0.00}$ & 23.30$_{\pm0.01}$ & 14.28$_{\pm0.01}$ & 23.75$_{\pm0.04}$ & 20.71$_{\pm0.01}$ & 17.53$_{\pm0.01}$ & 32.42$_{\pm0.05}$ & 21.17$_{\pm0.03}$ & 14.99$_{\pm0.02}$ & 25.97$_{\pm0.06}$ & 18.63$_{\pm0.04}$ \\
\midrule
\textbf{Ours} & EvoST & \best{9.48$_{\pm0.03}$} & \best{15.12$_{\pm0.04}$} & \best{21.77$_{\pm0.17}$} & \best{11.51$_{\pm0.20}$} & \best{17.87$_{\pm0.29}$} & \best{19.18$_{\pm0.35}$} & \best{13.45$_{\pm0.06}$} & \best{24.56$_{\pm0.09}$} & \best{19.85$_{\pm0.19}$} & \best{11.55$_{\pm0.04}$} & \best{19.57$_{\pm0.09}$} & \best{16.89$_{\pm0.07}$} \\
\bottomrule
\end{tabular}%
}
\end{table*}

\providecommand{\best}[1]{\cellcolor{yellow!35}\textbf{\textcolor{red}{#1}}}
\providecommand{\second}[1]{\cellcolor{black!10}\underline{#1}}
\begin{table*}[!t]
\centering
\caption{Average results on PEMS10--12 (mean$\pm$standard deviation, with the sample standard deviation over five seeds for EvoST).}
\label{tab:main_by_method_part3}
\scriptsize
\renewcommand{\arraystretch}{0.85}
\resizebox{\textwidth}{!}{%
\begin{tabular}{ll|ccc|ccc|ccc}
\toprule
\multirow{2}{*}{\textbf{Category}} & \multirow{2}{*}{\textbf{Baseline}} & \multicolumn{3}{c|}{\textbf{PEMS10}} & \multicolumn{3}{c|}{\textbf{PEMS11}} & \multicolumn{3}{c}{\textbf{PEMS12}} \\
\cmidrule(lr){3-5} \cmidrule(lr){6-8} \cmidrule(lr){9-11}
 & & \textbf{MAE} & \textbf{RMSE} & \textbf{MAPE (\%)} & \textbf{MAE} & \textbf{RMSE} & \textbf{MAPE (\%)} & \textbf{MAE} & \textbf{RMSE} & \textbf{MAPE (\%)} \\
\midrule
\multirow{4}{*}{\textbf{\shortstack[l]{Static\\STGNN\\Backbones}}} & DCRNN & 12.26$_{\pm0.16}$ & 20.13$_{\pm0.26}$ & 31.65$_{\pm0.35}$ & 19.69$_{\pm0.30}$ & 33.66$_{\pm0.34}$ & 31.28$_{\pm0.86}$ & 16.16$_{\pm0.17}$ & 28.46$_{\pm0.25}$ & 30.90$_{\pm0.78}$ \\
 & ASTGNN & 12.45$_{\pm0.05}$ & 20.80$_{\pm0.06}$ & 30.22$_{\pm0.32}$ & 20.00$_{\pm0.16}$ & 34.32$_{\pm0.27}$ & 26.69$_{\pm0.39}$ & 15.98$_{\pm0.10}$ & 28.49$_{\pm0.18}$ & 27.32$_{\pm0.53}$ \\
 & TGCN & 12.43$_{\pm0.08}$ & 21.49$_{\pm0.14}$ & 30.86$_{\pm0.22}$ & 21.35$_{\pm0.31}$ & 40.11$_{\pm0.53}$ & 30.37$_{\pm0.32}$ & 17.13$_{\pm0.08}$ & 33.43$_{\pm0.19}$ & 30.68$_{\pm0.38}$ \\
 & GWN & 12.54$_{\pm0.11}$ & 20.45$_{\pm0.13}$ & 30.65$_{\pm0.69}$ & 20.29$_{\pm0.44}$ & 34.35$_{\pm0.80}$ & 27.55$_{\pm0.81}$ & 16.38$_{\pm0.23}$ & 28.47$_{\pm0.35}$ & 27.32$_{\pm0.41}$ \\
\midrule
\multirow{4}{*}{\textbf{\shortstack[l]{Na\"ive \\Schemes}}} & Pretrain & 81.39$_{\pm3.16}$ & 84.52$_{\pm3.24}$ & 755.06$_{\pm28.01}$ & 85.43$_{\pm3.52}$ & 116.95$_{\pm3.38}$ & 82.12$_{\pm18.72}$ & 16.41$_{\pm0.21}$ & 27.96$_{\pm0.24}$ & 28.48$_{\pm0.81}$ \\
 & Retrain & 11.85$_{\pm0.04}$ & 19.66$_{\pm0.03}$ & 30.30$_{\pm0.22}$ & 18.77$_{\pm0.13}$ & 32.21$_{\pm0.15}$ & 27.67$_{\pm0.81}$ & 14.91$_{\pm0.05}$ & 26.54$_{\pm0.06}$ & 26.88$_{\pm0.52}$ \\
 & Online-NN & 12.35$_{\pm0.23}$ & 20.58$_{\pm0.45}$ & 31.79$_{\pm1.11}$ & 21.54$_{\pm1.59}$ & 36.45$_{\pm2.86}$ & 30.83$_{\pm1.25}$ & 18.04$_{\pm0.57}$ & 32.01$_{\pm1.14}$ & 29.42$_{\pm0.55}$ \\
 & Online-AN & 11.60$_{\pm0.09}$ & 19.22$_{\pm0.17}$ & 29.80$_{\pm0.34}$ & 18.29$_{\pm0.13}$ & 31.40$_{\pm0.13}$ & 25.39$_{\pm0.75}$ & 14.54$_{\pm0.06}$ & 25.97$_{\pm0.07}$ & 25.48$_{\pm0.28}$ \\
\midrule
\multirow{4}{*}{\textbf{\shortstack[l]{Evolving\\Graph}}} & TrafficStream & 11.93$_{\pm0.08}$ & 19.77$_{\pm0.14}$ & 32.18$_{\pm0.70}$ & 18.67$_{\pm0.35}$ & 31.90$_{\pm0.47}$ & 28.29$_{\pm1.63}$ & 14.74$_{\pm0.07}$ & 26.32$_{\pm0.10}$ & 26.88$_{\pm0.62}$ \\
 & PECPM & 12.61$_{\pm1.16}$ & 21.29$_{\pm2.68}$ & 30.41$_{\pm1.69}$ & 25.89$_{\pm2.53}$ & 44.81$_{\pm4.91}$ & 36.90$_{\pm6.66}$ & 16.18$_{\pm1.49}$ & 28.57$_{\pm3.19}$ & 29.30$_{\pm1.38}$ \\
 & STKEC & 11.92$_{\pm0.07}$ & 19.80$_{\pm0.11}$ & 31.06$_{\pm0.42}$ & 19.02$_{\pm0.23}$ & 32.38$_{\pm0.23}$ & 30.78$_{\pm1.62}$ & 14.71$_{\pm0.08}$ & 26.28$_{\pm0.12}$ & 26.66$_{\pm0.60}$ \\
 & EAC & 13.60$_{\pm1.76}$ & 21.08$_{\pm1.43}$ & 59.52$_{\pm40.27}$ & 25.76$_{\pm4.14}$ & 41.08$_{\pm5.52}$ & 47.34$_{\pm12.72}$ & 14.98$_{\pm0.30}$ & 26.18$_{\pm0.40}$ & 29.71$_{\pm2.65}$ \\
\midrule
\multirow{2}{*}{\textbf{\shortstack[l]{Retrieval\\TTC}}} & STRAP & 11.79$_{\pm0.09}$ & 19.58$_{\pm0.10}$ & 30.34$_{\pm0.58}$ & 18.60$_{\pm0.15}$ & 31.88$_{\pm0.20}$ & 28.85$_{\pm1.00}$ & 14.88$_{\pm0.14}$ & 26.48$_{\pm0.17}$ & 27.09$_{\pm1.22}$ \\
 & ST-TTC & 11.76$_{\pm0.04}$ & 19.53$_{\pm0.04}$ & 30.14$_{\pm0.22}$ & 18.52$_{\pm0.13}$ & 31.83$_{\pm0.15}$ & 27.53$_{\pm0.80}$ & 14.74$_{\pm0.05}$ & 26.28$_{\pm0.05}$ & 26.77$_{\pm0.51}$ \\
\midrule
\multirow{5}{*}{\textbf{\shortstack[l]{Static\\Forecasting\\Backbones}}} & STID & 12.53$_{\pm0.08}$ & 20.54$_{\pm0.09}$ & 33.11$_{\pm0.25}$ & 20.02$_{\pm0.32}$ & 33.89$_{\pm0.36}$ & 32.17$_{\pm1.75}$ & 16.07$_{\pm0.07}$ & 28.14$_{\pm0.19}$ & 33.77$_{\pm0.88}$ \\
 & STNorm & 13.75$_{\pm0.32}$ & 21.81$_{\pm0.40}$ & 41.57$_{\pm1.79}$ & 21.77$_{\pm0.34}$ & 35.62$_{\pm0.53}$ & 48.00$_{\pm2.68}$ & 19.49$_{\pm0.39}$ & 32.19$_{\pm0.51}$ & 58.93$_{\pm2.28}$ \\
 & iTrans & 11.83$_{\pm0.07}$ & 19.63$_{\pm0.10}$ & 30.28$_{\pm0.74}$ & 19.24$_{\pm0.28}$ & 32.79$_{\pm0.31}$ & 30.95$_{\pm2.77}$ & 15.27$_{\pm0.11}$ & 26.92$_{\pm0.15}$ & 29.32$_{\pm1.41}$ \\
 & DLinear & 12.86$_{\pm0.04}$ & 21.36$_{\pm0.05}$ & 31.07$_{\pm0.38}$ & 20.62$_{\pm0.12}$ & 35.52$_{\pm0.13}$ & 27.66$_{\pm0.61}$ & 16.23$_{\pm0.02}$ & 29.24$_{\pm0.05}$ & 27.74$_{\pm0.47}$ \\
 & STAEFormer & \second{10.34$_{\pm0.03}$} & \second{17.20$_{\pm0.05}$} & \second{27.83$_{\pm0.45}$} & \second{17.05$_{\pm0.12}$} & \second{28.82$_{\pm0.20}$} & \second{24.54$_{\pm0.69}$} & \second{12.91$_{\pm0.11}$} & \second{23.13$_{\pm0.12}$} & \second{23.94$_{\pm0.74}$} \\
\midrule
\multirow{6}{*}{\textbf{\shortstack[l]{Times-series\\Forecasting\\Foundation\\Model}}} & Chronos-2 (ZS) & 13.30$_{\pm0.00}$ & 22.54$_{\pm0.00}$ & 30.51$_{\pm0.00}$ & 20.04$_{\pm0.00}$ & 37.05$_{\pm0.00}$ & 25.36$_{\pm0.00}$ & 16.67$_{\pm0.00}$ & 30.85$_{\pm0.00}$ & 26.63$_{\pm0.00}$ \\
 & Chronos-2 (FT) & 11.85$_{\pm0.01}$ & 20.64$_{\pm0.04}$ & 27.84$_{\pm0.07}$ & 17.19$_{\pm0.02}$ & 32.41$_{\pm0.08}$ & 21.93$_{\pm0.04}$ & 14.66$_{\pm0.01}$ & 27.68$_{\pm0.07}$ & 22.84$_{\pm0.04}$ \\
 & TimesFM2.5 (ZS) & 13.63$_{\pm0.00}$ & 22.92$_{\pm0.00}$ & 31.47$_{\pm0.00}$ & 20.91$_{\pm0.00}$ & 38.14$_{\pm0.00}$ & 26.41$_{\pm0.00}$ & 17.48$_{\pm0.00}$ & 31.92$_{\pm0.00}$ & 27.93$_{\pm0.00}$ \\
 & TimesFM2.5 (FT) & 12.27$_{\pm0.01}$ & 21.11$_{\pm0.01}$ & 28.62$_{\pm0.03}$ & 18.32$_{\pm0.02}$ & 34.43$_{\pm0.07}$ & 23.11$_{\pm0.03}$ & 15.69$_{\pm0.02}$ & 29.74$_{\pm0.05}$ & 24.31$_{\pm0.02}$ \\
 & Moirai-2.0 (ZS) & 14.18$_{\pm0.00}$ & 23.64$_{\pm0.00}$ & 30.73$_{\pm0.00}$ & 21.24$_{\pm0.00}$ & 38.56$_{\pm0.00}$ & 25.78$_{\pm0.00}$ & 17.90$_{\pm0.00}$ & 32.32$_{\pm0.00}$ & 27.26$_{\pm0.00}$ \\
 & Moirai-2.0 (FT) & 12.43$_{\pm0.01}$ & 21.32$_{\pm0.03}$ & 29.22$_{\pm0.02}$ & 17.71$_{\pm0.01}$ & 32.78$_{\pm0.04}$ & 23.23$_{\pm0.03}$ & 15.32$_{\pm0.01}$ & 28.62$_{\pm0.02}$ & 24.38$_{\pm0.02}$ \\
\midrule
\textbf{Ours} & EvoST & \best{9.76$_{\pm0.03}$} & \best{16.30$_{\pm0.04}$} & \best{27.56$_{\pm0.12}$} & \best{13.70$_{\pm0.09}$} & \best{25.24$_{\pm0.12}$} & \best{22.27$_{\pm0.17}$} & \best{11.76$_{\pm0.06}$} & \best{21.36$_{\pm0.10}$} & \best{22.73$_{\pm0.21}$} \\
\bottomrule
\end{tabular}%
}
\end{table*}

\paragraph{Average performance on the remaining districts.}
\Cref{tab:main_by_method_part2,tab:main_by_method_part3} report PEMS05 to PEMS08 and PEMS10 to PEMS12. Together with \Cref{tab:main_by_method_part1}, EvoST has the lowest MAE, RMSE, and MAPE among the non-foundation methods in all 30 Avg dataset-metric comparisons, each strictly below every other non-foundation method. Averaged equally across the seven networks of these two tables, its MAE, RMSE, and MAPE are 10.5\%, 8.1\%, and 5.3\% lower than those of the STAEFormer row (13.1\%, 10.6\%, and 6.8\% across all ten networks). The STAEFormer row reports the benchmark's results, which differ from EvoST in input standardization and in some checkpoint selections (Section~\ref{exp}). \Cref{fig:annual_controls} isolates the effect of maintaining the same checkpoints.

\paragraph{Table conventions.}
In \Cref{tab:main_by_method_part2,tab:main_by_method_part3}, each row is one
baseline and each dataset reports MAE, RMSE, and MAPE in percent. Avg denotes
the mean of cumulative-horizon scores over prediction lengths 1--12. Entries
are mean$\pm$standard deviation (SD) where available. For EvoST, the SD is the
sample SD over five seeds. Deterministic zero-shot references are single-run and
shown with $\pm0.00$. Chronos-2 uses multivariate graph-grouped inference, with
per-year low-rank adaptation (LoRA) for its FT variant. Bold and
underlined entries mark the best and second-best non-foundation methods,
respectively. Foundation-model rows are excluded from highlighting.

\providecommand{\best}[1]{\cellcolor{yellow!35}\textbf{\textcolor{red}{#1}}}%
\providecommand{\second}[1]{\cellcolor{black!10}\underline{#1}}%
\begin{table*}[!h]
\centering
\caption{Per-horizon results, part 1/2, on PEMS03, PEMS04, PEMS05, PEMS06, and PEMS07 (mean$\pm$std over seeds per dataset). \textbf{Bold}: best, \underline{underline}: second best. EvoST is measured over five seeds (sample standard deviation). MAPE values are reported in percent. Chronos-2 (ZS) and Chronos-2 (FT) use multivariate graph-grouped inference, with per-year LoRA fine-tuning for FT. TSFM columns are excluded from highlighting. Compact headers denote zero-shot/fine-tuned pairs: Chro2-Z/F for Chronos-2, TimesF-Z/T for TimesFM2.5, Moira1-Z/T for Moirai-MoE, and Moira2-Z/T for Moirai-2.0. Their deterministic zero-shot (ZS) references are single runs, shown with $\pm0.00$ for visual consistency. Fine-tuned (FT) references are reported as mean$\pm$std over 3 seeds.}
\label{tab:main_part1}
\setlength{\tabcolsep}{3pt}
\renewcommand{\arraystretch}{0.95}
\resizebox{\textwidth}{!}{%
\begin{tabular}{cccccccccccccccccccccccccccccc}
\toprule
& & \multicolumn{4}{c|}{\textbf{Static STGNN Backbones}} & \multicolumn{4}{c|}{\textbf{Na\"ive Schemes}} & \multicolumn{4}{c|}{\textbf{Evolving Graph}} & \multicolumn{2}{c|}{\textbf{Retrieval TTC}} & \multicolumn{5}{c|}{\textbf{Static Forecasting Backbones}} & \multicolumn{8}{c|}{\textbf{TSFMs}} & \multicolumn{1}{c}{\textbf{Ours}} \\
\cmidrule(lr){3-6} \cmidrule(lr){7-10} \cmidrule(lr){11-14} \cmidrule(lr){15-16} \cmidrule(lr){17-21} \cmidrule(lr){22-29}
\textbf{Metric} & \textbf{Len} & \textbf{DCRNN} & \textbf{ASTGNN} & \textbf{TGCN} & \textbf{GWN} & \textbf{Pretrain} & \textbf{Retrain} & \textbf{OL-NN} & \textbf{OL-AN} & \textbf{TrafStm} & \textbf{PECPM} & \textbf{STKEC} & \textbf{EAC} & \textbf{STRAP} & \textbf{ST-TTC} & \textbf{STID} & \textbf{STNorm} & \textbf{iTrans} & \textbf{DLinear} & \textbf{STAE} & \textbf{Chro2-Z} & \textbf{Chro2-F} & \textbf{TimesF-Z} & \textbf{TimesF-T} & \textbf{Moira1-Z} & \textbf{Moira1-T} & \textbf{Moira2-Z} & \textbf{Moira2-T} & \textbf{EvoST} \\
\midrule
\multicolumn{30}{c}{\textbf{PEMS03}} \\
\midrule
\multirow{4}{*}{MAE} & 3 & 14.42$_{\pm0.21}$ & 13.70$_{\pm0.12}$ & 15.14$_{\pm0.10}$ & 14.22$_{\pm0.29}$ & 30.61$_{\pm1.22}$ & 13.09$_{\pm0.16}$ & 16.25$_{\pm0.19}$ & 12.76$_{\pm0.05}$ & 13.22$_{\pm0.11}$ & 13.88$_{\pm0.85}$ & 13.42$_{\pm0.27}$ & 14.18$_{\pm0.31}$ & 13.10$_{\pm0.12}$ & 12.89$_{\pm0.15}$ & 14.28$_{\pm0.15}$ & 16.46$_{\pm0.42}$ & 13.16$_{\pm0.13}$ & 14.17$_{\pm0.07}$ & \second{11.81$_{\pm0.11}$} & 12.53$_{\pm0.00}$ & 11.73$_{\pm0.01}$ & 13.20$_{\pm0.00}$ & 12.13$_{\pm0.01}$ & 14.50$_{\pm0.00}$ & 16.03$_{\pm0.64}$ & 13.66$_{\pm0.00}$ & 12.16$_{\pm0.00}$ & \best{10.40$_{\pm0.04}$} \\
\bottomrule
\end{tabular}%
}
\end{table*}

\providecommand{\best}[1]{\cellcolor{yellow!35}\textbf{\textcolor{red}{#1}}}%
\providecommand{\second}[1]{\cellcolor{black!10}\underline{#1}}%
\begin{table*}[!h]
\centering
\caption{Per-horizon results, part 2/2, on PEMS08, PEMS10, PEMS11, PEMS12, and TFNSW (mean$\pm$SD, with the sample SD over five seeds for EvoST).}
\label{tab:main_part2}
\setlength{\tabcolsep}{3pt}
\renewcommand{\arraystretch}{0.95}
\resizebox{\textwidth}{!}{%
\begin{tabular}{cccccccccccccccccccccccccccccc}
\toprule
& & \multicolumn{4}{c|}{\textbf{Static STGNN Backbones}} & \multicolumn{4}{c|}{\textbf{Na\"ive Schemes}} & \multicolumn{4}{c|}{\textbf{Evolving Graph}} & \multicolumn{2}{c|}{\textbf{Retrieval TTC}} & \multicolumn{5}{c|}{\textbf{Static Forecasting Backbones}} & \multicolumn{8}{c|}{\textbf{TSFMs}} & \multicolumn{1}{c}{\textbf{Ours}} \\
\cmidrule(lr){3-6} \cmidrule(lr){7-10} \cmidrule(lr){11-14} \cmidrule(lr){15-16} \cmidrule(lr){17-21} \cmidrule(lr){22-29}
\textbf{Metric} & \textbf{Len} & \textbf{DCRNN} & \textbf{ASTGNN} & \textbf{TGCN} & \textbf{GWN} & \textbf{Pretrain} & \textbf{Retrain} & \textbf{OL-NN} & \textbf{OL-AN} & \textbf{TrafStm} & \textbf{PECPM} & \textbf{STKEC} & \textbf{EAC} & \textbf{STRAP} & \textbf{ST-TTC} & \textbf{STID} & \textbf{STNorm} & \textbf{iTrans} & \textbf{DLinear} & \textbf{STAE} & \textbf{Chro2-Z} & \textbf{Chro2-F} & \textbf{TimesF-Z} & \textbf{TimesF-T} & \textbf{Moira1-Z} & \textbf{Moira1-T} & \textbf{Moira2-Z} & \textbf{Moira2-T} & \textbf{EvoST} \\
\midrule
\multicolumn{30}{c}{\textbf{PEMS08}} \\
\midrule
\multirow{4}{*}{MAE} & 3 & 14.86$_{\pm0.14}$ & 13.63$_{\pm0.05}$ & 15.01$_{\pm0.19}$ & 14.62$_{\pm0.19}$ & 52.44$_{\pm2.51}$ & 13.13$_{\pm0.06}$ & 16.14$_{\pm0.04}$ & 12.76$_{\pm0.09}$ & 12.85$_{\pm0.12}$ & 14.58$_{\pm0.83}$ & 13.32$_{\pm0.44}$ & 13.86$_{\pm0.28}$ & 12.98$_{\pm0.04}$ & 12.95$_{\pm0.06}$ & 14.23$_{\pm0.08}$ & 17.19$_{\pm0.35}$ & 13.42$_{\pm0.09}$ & 13.79$_{\pm0.05}$ & \second{11.99$_{\pm0.04}$} & 13.26$_{\pm0.00}$ & 12.46$_{\pm0.01}$ & 13.94$_{\pm0.00}$ & 12.77$_{\pm0.02}$ & 15.65$_{\pm0.00}$ & 17.49$_{\pm0.13}$ & 14.90$_{\pm0.00}$ & 13.02$_{\pm0.02}$ & \best{11.01$_{\pm0.03}$} \\
\bottomrule
\end{tabular}%
}
\end{table*}

\paragraph{Per-horizon performance and TSFM comparison.}
For compactness, the preceding tables average cumulative errors over prediction lengths 1 to 12. \Cref{tab:main_part1,tab:main_part2} report horizons 3, 6, and 12. EvoST has the lowest error among the non-foundation methods in all 120 reported cells (three metrics at horizons 3, 6, and 12 and Avg on ten networks). Averaged equally over the ten networks, its MAE is 14.0\%, 12.8\%, and 12.0\% lower than that of the STAEFormer column at horizons 3, 6, and 12, with the same caveat on this reported baseline as above.

Across the three horizons and Avg rows, EvoST also has lower error than the best TSFM column in 102 of the 120 cells, including all 80 MAE and RMSE cells and 27 of the 30 Avg cells. The 18 exceptions are MAPE cells: horizon 3 on all ten networks, horizon 6 on PEMS04, PEMS05, PEMS07, PEMS11, and PEMS12, and Avg on PEMS04, PEMS05, and PEMS11. Fine-tuning remains important for the foundation models: Chronos-2, TimesFM~2.5, and Moirai-2.0 improve over their zero-shot counterparts in all 30 Avg dataset-metric comparisons, with mean relative reductions of 13.5\%, 10.9\%, and 15.4\%, respectively.

\paragraph{Per-horizon table conventions.}
In \Cref{tab:main_part1,tab:main_part2}, EvoST is evaluated over five seeds
with the sample SD, and MAPE is reported in percent. Compact headers
pair zero-shot and fine-tuned variants: Chro2-Z/F, TimesF-Z/T, Moira1-Z/T,
and Moira2-Z/T. Deterministic zero-shot references are single-run and shown
with $\pm0.00$, whereas fine-tuned references report mean$\pm$SD over three
seeds. Chronos-2 uses multivariate graph-grouped inference and per-year LoRA
fine-tuning for its FT variant. TSFM columns are excluded from best/second-best
highlighting. Pretrain fixes the first-year model, Retrain starts from scratch each year,
Online-NN updates new sensors, and Online-AN updates all active sensors.

\subsubsection{Per-Step Results on All Ten Networks}
\Cref{per_horizon_all_app} extends \Cref{per_horizon} to all ten networks. The curves use per-step rather than cumulative errors and compare EvoST with the same checkpoints without adaptation (Frozen) and with the calibrator only (Frozen+RLS), over the same years and five seeds on each network.

 \begin{figure*}[!h]
    \centering
    \includegraphics[width=0.99\linewidth]{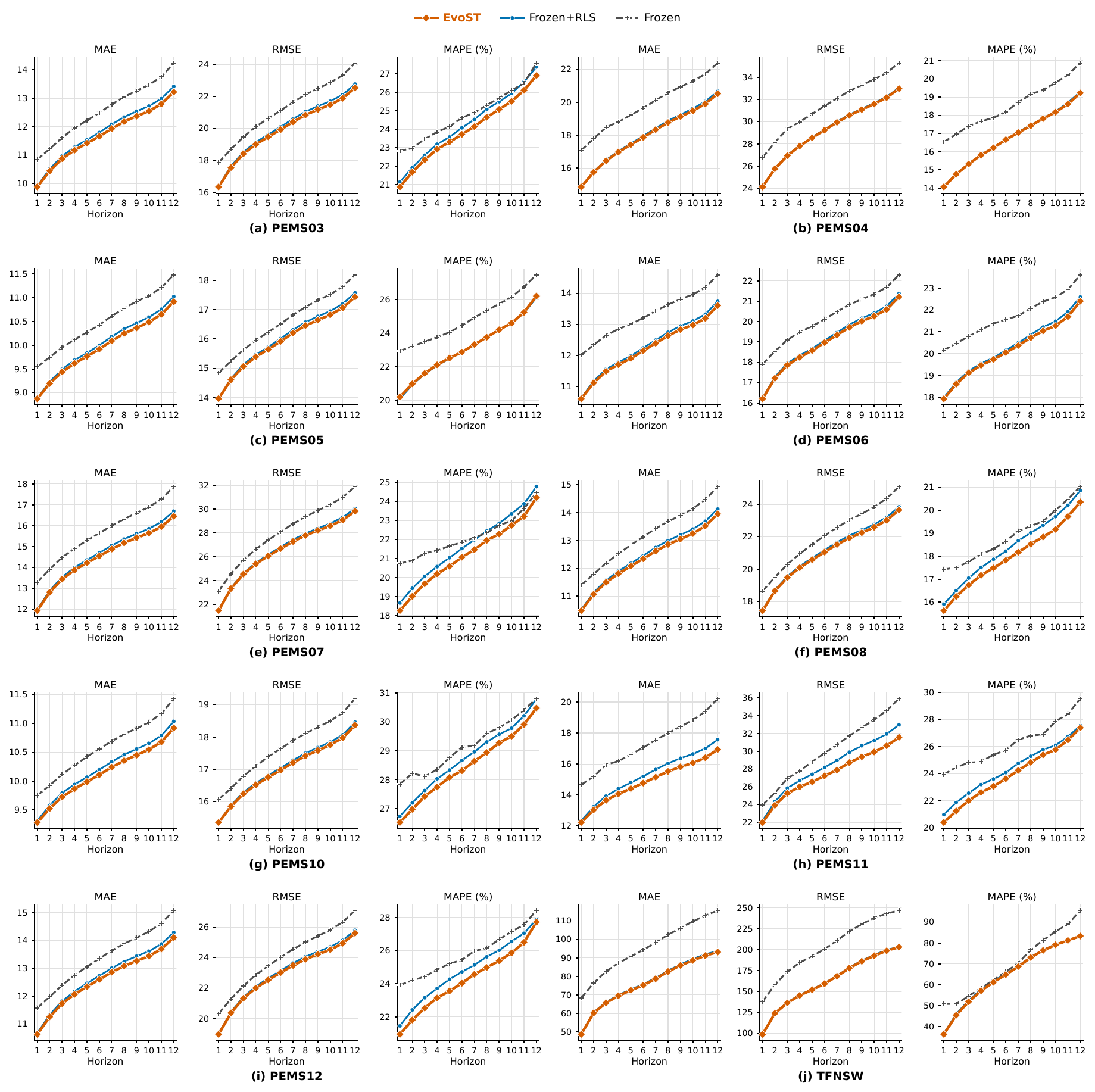}
    \caption{All-sensor per-step errors of EvoST and its matched controls on all ten networks: the same checkpoints with the calibrator only (Frozen+RLS) and without adaptation (Frozen), over the same years and five seeds. Lower is better.}
    \label{per_horizon_all_app}
    \Description{Thirty line plots compare MAE, RMSE, and MAPE of EvoST,
    Frozen+RLS, and Frozen at each of the 12 forecast steps for all sensors in
    nine EvoXXLTraffic districts and TFNSW.}
\end{figure*}

\paragraph{All sensors.}
In \Cref{per_horizon_all_app}, EvoST has lower error than Frozen in all $10\times3\times12=360$ combinations of network, metric, and step, and lower error than Frozen+RLS in 343. The 17 exceptions are MAPE at steps 1--7 and 9--12 on PEMS05 and at steps 4--6 on PEMS04, RMSE at the first step on PEMS04 and PEMS05, and MAPE at the last step on TFNSW. Averaged equally over networks and steps, EvoST lowers MAE, RMSE, and MAPE by 8.9\%, 7.2\%, and 6.3\% relative to Frozen and by 1.0\%, 0.8\%, and 1.2\% relative to Frozen+RLS.

\fi

\subsection{Annual Benchmark Protocol and Controls}
\label{app:annual}

\ifextended
\begin{figure*}[t]
    \centering
    \includegraphics[width=\textwidth]{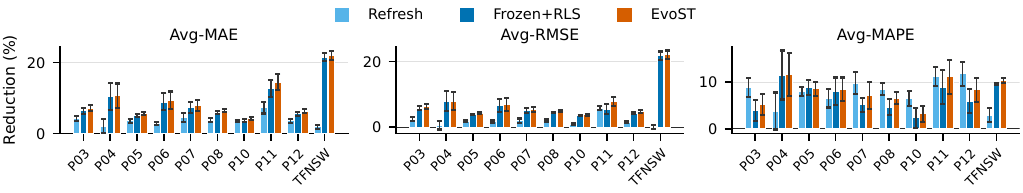}
\else
\begin{figure}[t]
    \centering
    \includegraphics[width=\linewidth]{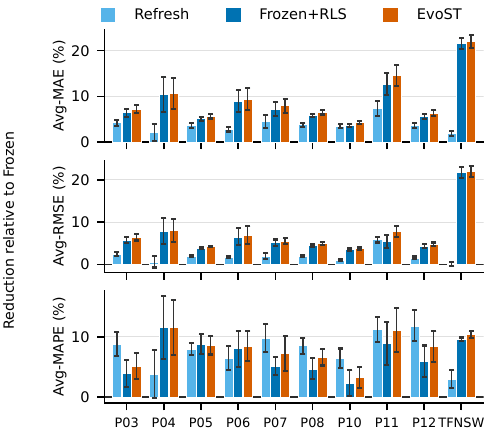}
\fi
    \caption{Matched controls on all ten networks: reduction in Avg-MAE, Avg-RMSE, and Avg-MAPE relative to the same per-year STAEFormer checkpoints without adaptation (Frozen), for refreshed identities only (Refresh), the calibrator only (Frozen+RLS), and both (EvoST). Bars show reductions of the five-seed means and whiskers the sample standard deviation of the per-seed reductions.}
    \label{fig:annual_controls}
    \Description{Three bar charts show ten networks for each metric. For Avg-MAE and Avg-RMSE, Frozen+RLS gives most of the reduction, and EvoST gives slightly larger reductions on every network. Refresh alone gives smaller reductions. For Avg-MAPE, Refresh alone exceeds EvoST on several networks.}
\ifextended
\end{figure*}
\else
\end{figure}
\fi

\ifextended
\begin{table*}[t]
\centering
\caption{Matched controls of the annual benchmark on all ten networks. Avg errors are reported for the same per-year STAEFormer checkpoints without adaptation (Frozen), with the calibrator only (Frozen+RLS), with refreshed identities only (Refresh), and with both (EvoST), as mean $\pm$ sample standard deviation over five seeds. MAPE is in percent.}
\label{tab:annual_controls}
\setlength{\tabcolsep}{2.2pt}
\scriptsize
\resizebox{\textwidth}{!}{%
\begin{tabular}{l|ccc|ccc|ccc|ccc}
\toprule
 & \multicolumn{3}{c|}{Frozen} & \multicolumn{3}{c|}{Frozen+RLS} & \multicolumn{3}{c|}{Refresh} & \multicolumn{3}{c}{EvoST} \\
Network & MAE & RMSE & MAPE & MAE & RMSE & MAPE & MAE & RMSE & MAPE & MAE & RMSE & MAPE \\
\midrule
PEMS03 & 11.76$_{\pm0.15}$ & 19.72$_{\pm0.19}$ & 23.75$_{\pm0.73}$ & 11.01$_{\pm0.04}$ & 18.58$_{\pm0.06}$ & 22.82$_{\pm0.26}$ & 11.26$_{\pm0.09}$ & 19.24$_{\pm0.15}$ & 21.66$_{\pm0.24}$ & 10.92$_{\pm0.05}$ & 18.47$_{\pm0.07}$ & 22.52$_{\pm0.21}$ \\
PEMS04 & 18.57$_{\pm0.91}$ & 29.49$_{\pm1.11}$ & 17.55$_{\pm1.20}$ & 16.63$_{\pm0.15}$ & 27.17$_{\pm0.18}$ & 15.53$_{\pm0.20}$ & 18.18$_{\pm1.22}$ & 29.32$_{\pm1.47}$ & 16.88$_{\pm1.86}$ & 16.58$_{\pm0.21}$ & 27.14$_{\pm0.26}$ & 15.52$_{\pm0.23}$ \\
PEMS05 & 10.05$_{\pm0.04}$ & 15.79$_{\pm0.04}$ & 23.81$_{\pm0.50}$ & 9.53$_{\pm0.03}$ & 15.17$_{\pm0.05}$ & 21.72$_{\pm0.15}$ & 9.68$_{\pm0.05}$ & 15.49$_{\pm0.06}$ & 21.90$_{\pm0.41}$ & 9.48$_{\pm0.03}$ & 15.12$_{\pm0.04}$ & 21.77$_{\pm0.17}$ \\
PEMS06 & 12.70$_{\pm0.50}$ & 19.21$_{\pm0.63}$ & 20.95$_{\pm0.89}$ & 11.57$_{\pm0.20}$ & 17.96$_{\pm0.29}$ & 19.28$_{\pm0.29}$ & 12.35$_{\pm0.49}$ & 18.87$_{\pm0.64}$ & 19.59$_{\pm0.58}$ & 11.51$_{\pm0.20}$ & 17.87$_{\pm0.29}$ & 19.18$_{\pm0.35}$ \\
PEMS07 & 14.62$_{\pm0.31}$ & 25.98$_{\pm0.30}$ & 21.39$_{\pm0.67}$ & 13.56$_{\pm0.07}$ & 24.65$_{\pm0.09}$ & 20.29$_{\pm0.33}$ & 13.96$_{\pm0.11}$ & 25.47$_{\pm0.13}$ & 19.30$_{\pm0.16}$ & 13.45$_{\pm0.06}$ & 24.56$_{\pm0.09}$ & 19.85$_{\pm0.19}$ \\
PEMS08 & 12.35$_{\pm0.07}$ & 20.58$_{\pm0.10}$ & 18.08$_{\pm0.28}$ & 11.63$_{\pm0.04}$ & 19.66$_{\pm0.08}$ & 17.24$_{\pm0.10}$ & 11.88$_{\pm0.04}$ & 20.17$_{\pm0.11}$ & 16.54$_{\pm0.11}$ & 11.55$_{\pm0.04}$ & 19.57$_{\pm0.09}$ & 16.89$_{\pm0.07}$ \\
PEMS10 & 10.19$_{\pm0.06}$ & 16.94$_{\pm0.07}$ & 28.49$_{\pm0.53}$ & 9.82$_{\pm0.03}$ & 16.34$_{\pm0.04}$ & 27.81$_{\pm0.15}$ & 9.83$_{\pm0.04}$ & 16.78$_{\pm0.05}$ & 26.64$_{\pm0.12}$ & 9.76$_{\pm0.03}$ & 16.30$_{\pm0.04}$ & 27.56$_{\pm0.12}$ \\
PEMS11 & 16.04$_{\pm0.46}$ & 27.38$_{\pm0.45}$ & 25.05$_{\pm0.97}$ & 14.00$_{\pm0.08}$ & 25.89$_{\pm0.12}$ & 22.81$_{\pm0.23}$ & 14.87$_{\pm0.36}$ & 25.78$_{\pm0.40}$ & 22.24$_{\pm0.43}$ & 13.70$_{\pm0.09}$ & 25.24$_{\pm0.12}$ & 22.27$_{\pm0.17}$ \\
PEMS12 & 12.54$_{\pm0.14}$ & 22.42$_{\pm0.20}$ & 24.80$_{\pm0.84}$ & 11.85$_{\pm0.06}$ & 21.45$_{\pm0.10}$ & 23.32$_{\pm0.27}$ & 12.10$_{\pm0.09}$ & 22.08$_{\pm0.16}$ & 21.87$_{\pm0.20}$ & 11.76$_{\pm0.06}$ & 21.36$_{\pm0.10}$ & 22.73$_{\pm0.21}$ \\
TFNSW & 83.54$_{\pm1.68}$ & 176.72$_{\pm3.38}$ & 59.00$_{\pm0.34}$ & 65.54$_{\pm0.35}$ & 138.29$_{\pm0.53}$ & 53.33$_{\pm0.21}$ & 81.93$_{\pm1.37}$ & 176.64$_{\pm3.43}$ & 57.26$_{\pm0.98}$ & 65.16$_{\pm0.34}$ & 137.77$_{\pm0.59}$ & 52.89$_{\pm0.29}$ \\
\bottomrule
\end{tabular}}
\end{table*}

\fi

\paragraph{Protocol.}
Identity fitting uses at most 16 epochs and patience 3 per candidate learning
rate and samples every second window for fitting and holdout (every fourth
above 1,600 sensors). TFNSW in 2026 has only 1,704 training and 552
validation windows: the identity fit holds out the final 512 window starts
(596 fitting and 256 holdout windows after sampling), and the calibrators are
initialized on all 1,704 training windows, with settings selected on all 552
validation windows. In PEMS06 in 2005, the fitting windows contain no valid
target, so the five seeds keep the checkpoint identities and no calibrator can
be fitted: for this year only, the calibrators use the standard deviation of
the year's training inputs as $s$ and start every selection and test stream
from zero residual coefficients, then update on released labels as usual.
Refresh therefore coincides with Frozen and EvoST with Frozen+RLS. The lowest candidate learning rate was selected
in 1,100 of the other 1,160 runs (337 of 350 on PEMS03, PEMS04, and TFNSW).
The selected candidate ran all 16 epochs in 82 runs (47 on TFNSW), with the
16th as its best epoch in 30 (19 on TFNSW). The 1,165 runs took 100.3 H100
GPU-hours, including candidate fits, calibration selection, and diagnostic
replays.

\paragraph{Matched controls.}
The four streams in \Cref{fig:annual_controls} share the same per-year
checkpoints, training-split input normalization, windows, and seeds.
Their mean reductions relative to Frozen over the ten networks are shown in
\Cref{fig:ablation}(a)\extref{, and their absolute errors in
\Cref{tab:annual_controls}}{}. On average over the ten networks, EvoST's
Avg-MAE, Avg-RMSE, and Avg-MAPE are 0.8\%, 0.6\%, and 1.2\% below those of
Frozen+RLS. EvoST's Avg-MAE is lower on every network, by 0.3\% on PEMS04 to 2.1\% on
PEMS11, with lower cross-year means in all five paired seeds except on PEMS04
(three). On PEMS04, its Avg-RMSE and Avg-MAPE are lower in only two of the
five seeds. On PEMS05, its Avg-MAPE is 0.2\% higher. Calibration lowers MAE
and RMSE of both the frozen and the refreshed model, and MAPE of the frozen
model, on every network. For the refreshed model, it raises MAPE on six
networks (PEMS03, PEMS07, PEMS08, PEMS10, PEMS11, and PEMS12) and lowers it on
the other four.

\ifextended\else
\PerStepFigure{t}{1}{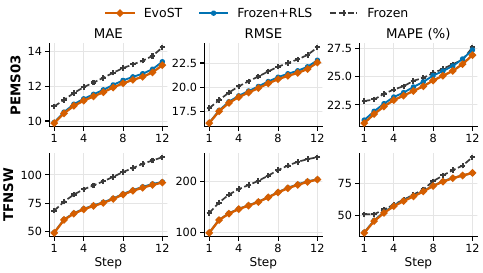}
\fi

\subsection{EvoST Implementation and Feedback Protocol}
\label{app:evost_implementation}

\ifextended\else
\EvoSTAlgorithm
\fi

\paragraph{Identity parameters and fitting.}
The identity blocks have shapes $12\times N_y\times80$ for STAEFormer,
$N_y\times32$ for STID, and two $N_y\times10$ tables for Graph WaveNet.
The main STAEFormer maintenance comparison uses the latest 2,048 training
windows, reserves the final quarter for early stopping, and selects the
identity learning rate from $\{10^{-3},3\times10^{-3},10^{-2}\}$. Arrival
experiments use a fixed rate of $3\times10^{-3}$. These fits use at most 16
epochs and patience 3.

\paragraph{Maintenance controls.}
The fitted variants shown in \Cref{fig:maintenance} use the same recent windows
and holdout as identity refresh and select their learning rates on that
holdout.
Head and Prompt select from $\{3\times10^{-3},10^{-2},3\times10^{-2}\}$ and
Shared from $\{10^{-4},3\times10^{-4},10^{-3}\}$. Full FT selects the rate of
the shared weights from $\{3\times10^{-5},10^{-4},3\times10^{-4}\}$ and uses
the identity rate selected for EvoST in the same run. EvoST+Prompt keeps that
identity rate and selects the prompt rate. These fits also use at most 16
epochs and patience 3. Scratch starts from random initialization with
target-year training statistics, selects from
$\{3\times10^{-4},10^{-3},3\times10^{-3}\}$, and uses at most 100 epochs with
patience 5. Its selected fits ran at most 41 epochs. The lowest rate of the
grid was selected in 27 of 36 runs for EvoST, 20 for Full FT, 21 for Shared,
and 23 for Scratch. Such lower-bound selections make the comparisons
conditional on the reported grids and fitting budgets.
The prompt is a rank-6 EAC-style input prompt $\mathbf U\mathbf V$ with
$\mathbf U\in\mathbb R^{N_y\times6}$ and $\mathbf V\in\mathbb R^{6\times12}$,
that is, $6N_y+72$ trainable values (median 8,856, against a median of
1,405,440 identity values). It does not reproduce the full EAC procedure.

\ifextended
\paragraph{Known and unseen sensors.}
On the main panel, 3,612 of the 18,996 sensors over the twelve settings (19.0\% overall,
5.4--46.0\% per setting) are absent from the five-year-old checkpoint. Old and
the fitted variants in \Cref{tab:known_unseen} start them from the checkpoint's
mean identity. \Cref{tab:known_unseen} splits \Cref{fig:maintenance} by sensor
group, with the same reference, averaging, and intervals.

\begin{table*}[t]
  \caption{Parameter maintenance of five-year-old STAEFormer checkpoints by sensor group. Known sensors are in the old
  checkpoint. Unseen ones are absent from it and start from its mean identity. Calibrated test H12 MAE ratios to the
  calibrated target-year checkpoint on the same sensors are geometric means over twelve settings after averaging three seeds,
  with 95\% district-bootstrap intervals. Counts sum over the settings. Recovery is the share of the Old gap closed by
  identity refresh (EvoST).}
  \label{tab:known_unseen}
  \centering
  \begin{tabular}{lrccccr}
    \toprule
    Sensors & Count & Old & Shared & EvoST & Full FT & Recovery \\
    \midrule
    Known & 15,384 & 1.169 [1.157, 1.182] & 1.067 [1.055, 1.079] & 1.026 [1.015, 1.037] & 1.018 [1.010, 1.028] & 85\% \\
    Unseen & 3,612 & 1.226 [1.207, 1.247] & 1.128 [1.114, 1.141] & 1.027 [1.017, 1.038] & 1.022 [1.015, 1.028] & 88\% \\
    All & 18,996 & 1.177 [1.166, 1.187] & 1.078 [1.068, 1.089] & 1.025 [1.015, 1.037] & 1.018 [1.010, 1.027] & 86\% \\
    \bottomrule
  \end{tabular}
\end{table*}
\fi

\begin{figure}[htbp]
    \centering
    \includegraphics[width=\linewidth]{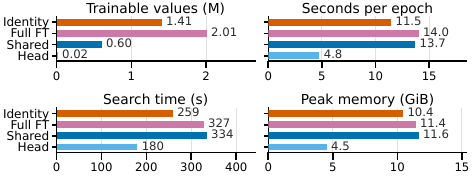}
    \caption{Median fitting costs over the 36 H100 runs of the primary STAEFormer maintenance experiment: identity refresh (Identity), full fine-tuning (Full FT), shared weights (Shared), and output head (Head).}
    \label{fig:cost}
    \Description{Four horizontal bar charts compare identity refresh, full fine-tuning, shared-weight tuning, and output-head tuning by trainable values, seconds per epoch, search time, and peak memory.}
\end{figure}

\paragraph{Fitting cost.}
\Cref{fig:cost} reports medians over the 36 H100 runs of the main maintenance
comparison. Identity refresh updates 960 values per sensor. Seconds per epoch
average the training-epoch times within each candidate and then over the
three candidates, excluding holdout evaluation. Search times sum each arm's
three candidate fits and exclude calibration and, for Full FT, the preceding
identity-rate search. Memory is the peak process allocation during fitting,
including resident states and caches.

\paragraph{Checkpoint age.}
The checkpoint-age study uses the six main-panel districts with five-year-old
checkpoints in 2015, 2020, and 2025, ten-year-old ones in 2020 and
2025, and fifteen-year-old ones in 2025 (18, 12, and 6 settings).

\paragraph{Backbone and age extensions.}
For the checkpoint-age, STID, and time-input experiments, identities are
refitted with a fixed learning rate of $3\times10^{-3}$, at most 16 epochs, and
patience 3. Graph WaveNet selects its rate from
$\{10^{-3},3\times10^{-3},10^{-2},3\times10^{-2}\}$ on the same fitting
holdout and budget. STID and Graph WaveNet are the benchmark implementations,
retrained per year with checkpoint seeds 42--44. Graph WaveNet is the
benchmark's compact two-layer version, which averages the given and the
learned adjacency. On the nine PeMS districts in 2025, the uncalibrated test
H12 MAE of the retrained models is 0.961 (STID) and 1.006 (Graph WaveNet)
times that of the benchmark's own checkpoints. With calibration, the
same-year STID and Graph WaveNet have 26.9\% and 22.9\% higher error than
STAEFormer on the sixteen settings of \Cref{tab:backbones}, so the ratios there measure
reuse within each backbone, not an accuracy ranking. The time-input models
add time-of-day and day-of-week embeddings, initialized from
$\mathcal N(0,0.02^2)$ by a rule fixed on development data. These models have width 152,
compared with 104 without time inputs. The shared time embeddings stay frozen during
refresh.

\paragraph{Retraining checks.}
The original check of the Graph WaveNet retraining flagged 11 of 90 models
whose training-mode validation error was not below that of the last-value
forecast. An evaluation-mode check found all 90 below it (largest ratio
0.976). Excluding the six affected settings, as the original rule
prescribed, gives 1.098 for the old checkpoint and 1.005 after refresh (95\%
recovery), against 1.091 and 1.011 with all sixteen. One time-input model diverged
(PEMS06/2025, checkpoint seed 43). We keep its best checkpoint (epoch 2 of
8). Excluding PEMS06/2025 afterwards gives 15.8\% ageing and
1.027 after refresh (83\% recovery), against 14.0\% and 1.014 with all
twelve settings.

\ifextended\else
\ArrivalFigure{0.88}
\fi

\paragraph{Arrival history and donors.}
The arrival experiment provides 64 post-arrival fitting windows before the
stream. Matching uses finite raw readings, including zeros. A donor must have
observations on at least $20\%$ of the available times and share finite
observations with the incoming sensor on at least $20\%$. If fewer than $k$
donors qualify, all $k$ are sampled without replacement from the established
sensors with a seed fixed for the run. At zero-based stream index
$t_\star=1{,}024$, with $H=12$, records $0,\ldots,1{,}012$ are eligible. These
1,013 records and the last 11 eligible post-arrival prestream windows supply
the $W=1{,}024$ refitting windows.

\paragraph{Arrival refits.}
The new-only refit fits the new identities and rebuilds only their
calibrators. The all-identity refit fits every identity and rebuilds every
calibrator. The established-only backbone reference refreshes identities
before the stream and still selects calibration hyperparameters on all-node
validation error, including last-value forecasts for the new nodes.
Established-sensor results include only established-node errors. We specified
the all-identity refit after the new-only refit left an established-sensor
gap and evaluated it on the same region-year settings. Before this
evaluation, we required the upper 95\% district-bootstrap bound to be at most
1.01 for established sensors relative to the established-only backbone
reference and for new sensors relative to the new-only refit. The bounds were
1.009 and 1.002. Without output calibration, the new-sensor MAE ratio to the
uncalibrated target-year model is 0.994 (95\% CI 0.973--1.012). Rebuilding
the new-sensor calibrators on cross-fitted forecasts did not help (ratio 1.004
to the new-only refit).

\paragraph{Ablation and sensitivity figures.}
In \Cref{fig:sensitivity}(a, b), colored curves average errors within each network
and then across the ten networks. In panel (b), grey lines show individual networks. In
\Cref{fig:ablation}(b, c) and \Cref{fig:sensitivity}(c, d), ratios are
geometric means over settings (twelve each, except for the checkpoint ages)
of seed-averaged test H12 MAE relative to the target-year checkpoint, both
calibrated or both not. For the history budgets, the target-year reference is
calibrated at the same budget. Whiskers are 95\% district-bootstrap intervals
(none for dashed curves). The budget and age panels fix the identity learning
rate at $3\times10^{-3}$. Percentages in the age panel are the share of the
gap recovered.

\paragraph{History budgets.}
With a budget of $b$ fitting windows, the latest $b$ training windows both
refresh the identities and initialize the calibrators. The final quarter is
the identity holdout, and fitting uses every second window (every fourth above
1,600 sensors). The identities stay fixed while the calibrators keep absorbing
released labels, so the budget limits the history available at the start, not
the labels used during the stream. All budgets use at most 16 epochs and
patience 3 with the same number of windows per epoch, so the number of steps
depends on early stopping. Calibration follows the protocol below.

\paragraph{Case study.}
\Cref{fig:case} uses evaluation seed 51 and the 2015 checkpoint. Panel (a)
uses the original maintenance evaluation and includes all 2,171 sensors with
valid test targets (249 have none). Panel (b) records a separate run of the
same protocol with the same seed and fixed sensors. All forecasts are
calibrated. Panel (a) uses MAE over steps 1--12, and panel (b) shows forecasts at step 12. The p50 and
p90 sensors have the nearest-rank median and 90th-percentile relative MAE
reductions among the 1,126 known sensors with at least 95\% valid test
targets, and the unseen sensor has the highest validity, ties broken by
sensor ID. These choices were fixed before the run of (b). The day shown is
the first complete weekday by 60-minute target time. Differences in
\Cref{sec:case} are given in the order p50, p90, unseen.

\paragraph{Calibration and evaluation.}
We use $L=H=12$, giving $D=29$, and update RLS every $\delta=64$ windows. The
history slope is fitted against centered sample indices. The recursion uses
float64 arithmetic and symmetrizes $\mathbf P_n$ after each update. Each
calibrator is initialized by ridge regression (\cref{eq:ridge_initialization})
on the training tail, which holds the latest 2,048 training windows (all of them if fewer, or
$b$ in the history-budget study). The scale $s$ is the standard deviation of the
valid targets in these windows. Starting from this fit, $\rho\in\{1,100\}$ and
$\lambda\in\{1,0.999,0.995,0.99\}$ are selected for each branch by streaming
up to the first 1,024 validation windows ($b/2$ in the history-budget study).
A new fit on the same windows then streams validation and test
chronologically in the maintenance experiments, and only the test portion in
the annual benchmark. Only test errors are reported. PEMS06 in 2005, whose
training tail has no valid target, follows the fallback in \Cref{app:annual}. The arrival
experiments compute $s$ from all sensors' valid targets in the tail before
applying the arrival masks, which restrict donor observations and
sensor-specific fitting records, so $s$ includes targets that new sensors
record before their arrival. Restricted to the targets the masks allow, $s$
would change by $-3.4\%$ to $+8.9\%$ across the twelve settings. Since $s$
only rescales the base-forecast features and the residual targets, this
equals multiplying the ridge penalty on the coefficients of the twelve scaled
base-forecast features by 0.93--1.19.
In the two settings with the largest scale changes (PEMS03 in 2025 and
PEMS06 in 2020), restricting $s$ preserves the selected calibration settings.
The per-setting point-estimate ratios of seed-mean test H12 MAE underlying
\Cref{fig:arrival} and the arrival-refit comparisons change by at most 0.034\%.
Replacing only these two settings and keeping the other ten fixed changes
the twelve-setting geometric means by less than 0.01\%.

\end{document}